\pdfoutput=1

\documentclass[11pt]{article}

\usepackage{acl}
\usepackage{times}
\usepackage{latexsym}

\usepackage[T1]{fontenc}

\usepackage[utf8]{inputenc}

\usepackage{microtype}

\usepackage{amsmath}
\usepackage{amssymb}
\usepackage{mathtools}
\usepackage{algorithm}
\usepackage{algorithmic}
\usepackage{adjustbox}
\usepackage{hyperref}
\usepackage{inconsolata}

\usepackage{graphicx}
\usepackage{xcolor}
\usepackage{tabularx}
\usepackage[lofdepth,lotdepth]{subfig}
\newcommand{\jaakko}[1]{\textcolor{black}{#1}}

\title{Constrained Non-negative Matrix Factorization for Guided Topic Modeling of Minority Topics}

\author{
  Seyedeh Fatemeh Ebrahimi \and Jaakko Peltonen\textsuperscript{\dag} \\
  Faculty of Information Technology and Communication Sciences \\
  Tampere University \\
  \href{mailto:seyedeh.ebrahimi@tuni.fi}{seyedeh.ebrahimi@tuni.fi},\ 
  \href{mailto:jaakko.peltonen@tuni.fi}{jaakko.peltonen@tuni.fi}
}

\begin{document}
\maketitle
\begin{abstract}
Topic models often fail to capture low-prevalence, domain-critical themes—so-called \textit{minority topics}—such as mental health themes in online comments. 
\jaakko{While some 
existing methods} 
\jaakko{can} 
incorporate domain knowledge such as expected topical content, 
\jaakko{methods} allowing guidance may require overly detailed expected topics, hindering the discovery of topic divisions and variation. We propose a topic modeling solution via a specially constrained NMF. We incorporate a seed word list characterizing minority content of interest, but we do not require experts to pre-specify their division across minority topics. Through prevalence constraints on minority topics and seed word content across topics, we learn distinct data-driven minority topics as well as majority topics. The constrained NMF is fitted via Karush-Kuhn-Tucker (KKT) conditions with multiplicative updates. We outperform several baselines on synthetic data in terms of topic purity, normalized mutual information, and also evaluate topic quality using Jensen-Shannon divergence (JSD). We conduct a case study on YouTube vlog comments, analyzing viewer discussion of mental health content; our model successfully identifies and reveals this 
domain-relevant 
minority 
content.
\end{abstract}

\begingroup
\renewcommand\thefootnote{\dag}
\footnotetext{Equal contribution.}
\endgroup

\section{Introduction}
A central problem in many data analysis domains is identifying and extracting both dominant and minority themes from extensive corpora \cite{Jagarlamudi2012}. Topic modeling is a well-established task proposed to discover latent themes from a collection of texts based on word occurrence \cite{Srivastava2017AutoencodingVI, Egger2022ATM, Wu2024ASO}, so that each topic represents a theme by grouping together related words. 
Minority topics are generally defined as 
\jaakko{themes that have low prevalence} both corpus-wide and within individual documents, \jaakko{and}
we specifically focus on \jaakko{\emph{domain-relevant minority content}}.
An example is mental health discussion in YouTube comments: it is rare compared to other discussion, and often mentioned only briefly within a comment. 

Minority topics are crucial for understanding niche but meaningful content—especially in domains like mental health—but tend to be overlooked by state-of-the-art (SOTA) models. 

Firstly, conventional methods adopt probabilistic approaches such as Latent Dirichlet Allocation (LDA; \citealt{10.5555/944919.944937}), and others such as non-negative matrix factorization (NMF; \citealt{leelda2000algorithms}) which has been extended to several scenarios, as well as clustering-based techniques \cite{chen2019experimental}. Neural and LLM-based topic models (e.g., Top2Vec~\cite{Angelov2020Top2VecDR}, BERTopic~\cite{grootendorst2022bertopic}, FASTopic~\cite{wu2024fastopic},~\cite{bianchi-etal-2021a-pre, bianchi-etal-2021b-cross}), and embedding-clustering approaches~\cite{sia-etal-2020-tired} propose contextual representations, but still struggle to identify minority themes due to dominance of frequent topics. 

Secondly, most topic models also lack the flexibility to incorporate domain knowledge, such as expert expectations about content.
Methods such as Anchored Correlation Explanation \cite{Gallagher2016AnchoredCE} that do allow expert guidance typically require detailed specifications of expected topics \cite{Steeg2014DiscoveringSI}. An expert may not be able to predefine them, and relying on them can limit discovery of variations within topics. 
Other models also employ guided \cite{9413656} or semi-supervised approaches \cite{Lee2010SemiSupervisedNM,  8820170, Lindstrom2022ContinuousSN} by incorporating prior knowledge to guide the model toward finding desired topics in various fashions \cite{a15050136}. 
However, 
such guided topic models often struggle to detect such low-prevalence themes, and many require strong assumptions, hyperparameter tuning, or rigid supervision, limiting their ability to generalize to subtle or unexpected variations.


\textbf{Our solution.}
To address the issues we propose a novel topic model using a specially constrained NMF. Our method integrates soft prevalence constraints and a unified seed word list, without requiring topic-specific supervision. \jaakko{We set} inequality constraints on topic distributions \jaakko{in}
documents and word distributions within topics.  
The model is optimized to minimize a generalized Kullback-Leibler divergence reconstruction \jaakko{error} 
under the constraints, using KKT conditions \cite{Lange2013,ghojogh2021kkt}, 
\jaakko{yielding}
multiplicative updates.
This 
\jaakko{lets us}
distinguish data-driven topics in a 
\jaakko{nuanced} way, ensuring minority themes are well represented in addition to majority ones.

\jaakko{Crucially, our model aims} to discover minority topics without requiring 
\jaakko{them to be dominant}
in the corpus 
\jaakko{and without requiring seed words to be prominent}
within such topics.
\jaakko{We} do not enforce presence of 
minority content in all topics, nor do we 
\jaakko{maximize} prevalence of minority content or prominence of seed words, as such approaches could distort their modeling.
Rather, by mild constraints on prevalence of minority topics \jaakko{and} 
distribution of seed word content, we 
\jaakko{let the model}
learn distinct, data-driven minority and majority topics. 

Unlike 
\jaakko{models} that force seed alignment or rely heavily on neural decoding~\cite{lin2023enhancing}, and models that require exact specified guidance and prior domain knowledge, our approach not only enhances representation of minority themes but also enables 
\jaakko{flexible} topic discovery without imposing rigid prior structures on the data. 
This can be crucial for analyzing 
\jaakko{data}
where subtle variations in themes 
\jaakko{are}
key to understanding the domain. 

\textbf{Our key contributions are:}
\textbf{1.}  We target underrepresented topics: instead of enforcing guidance on all topics, we identify a subset under guidance constraints while leaving others unconstrained for flexibility.
\textbf{2.}  We incorporate domain knowledge without overspecification: our guidance does not require preexisting knowledge of topic divisions.
\textbf{3.}  We apply soft prevalence constraints to avoid overfitting to seed words, enabling balanced topic emergence.
\textbf{4.}  Our model is grounded in constrained NMF, and optimized via KKT optimization with multiplicative updates.
\textbf{5.}  Experiments show improvements over several baselines on synthetic data and extract high-quality domain-relevant topics in a case study on real-world mental health data.

Next, Section \ref{sec:related_work} discusses the related work. Sections \ref{sec:method}, and  \ref{sec:optimization} detail our method and its optimization. Section \ref{sec:experiments} details comparison experiments and a mental health case study. Section \ref{sec:results} gives results and \ref{sec:discovered_topics} findings; Section \ref{sec:conclusions} concludes.

\section{Related Work}
\label{sec:related_work}

We review baseline models and their variants \cite{zhao2021topic}, 
existing 
\jaakko{NMF} and semi-supervised NMF models \cite{carbonetto2021non, Lindstrom2022ContinuousSN}, and other supervised models. 

Among probabilistic topic models LDA \cite{10.5555/944919.944937} has been widely used \cite{chen2019experimental}. LDA, LSA, and PLSA~\cite{10.1145/312624.312649, Albalawi2020UsingTM} are probabilistic models using Bayesian graphical structures with topics as latent variables. 
The methods 
prioritize discovering the most common patterns 
\jaakko{over}
documents as latent themes, 
\jaakko{but}
may struggle to represent less frequent trends \cite{10.1007/978-3-031-66336-9_26}. 
LDA models documents 
\jaakko{as}
\jaakko{bags-of-words} (BOW)
\jaakko{with word counts}
drawn from topic-specific word distributions~\cite{10.5555/944919.944937}. This can overemphasize frequent terms, and 
\jaakko{reliance} on Gibbs sampling degrades performance on short texts with sparse co-occurrence~\cite{chen2019experimental}. In contrast, NMF factorizes any non-negative matrix (e.g., TF-IDF) into interpretable components, and has proven effective in unsupervised clustering tasks~\cite{chen2019experimental, obadimu2019identifying, carbonetto2021non}. 
We 
\jaakko{use}
NMF as \jaakko{a} foundation of our method due to its flexibility 
\jaakko{but}
our constrained formulation is 
\jaakko{extendable} to LDA-style count models too.

Due to its simplicity and effectiveness, NMF has become influential in data mining \cite{zhang2012nonnegative}. While NMF can outperform LDA, yielding higher-quality topics on short-text datasets, traditional NMF shows limited effectiveness in discovering expected topics and often overlooks crucial minority content in document collections 
\cite{chen2019experimental, Egger2022ATM} and other downstream tasks \cite{9413656}. Unsupervised NMF approaches may learn meaningless or biased topics and often suffer from redundancy particularly when the data set is biased toward a set of features \cite{a15050136, NIPS2009_f92586a2, Jagarlamudi2012, 9413656}.
To address the limitations, researchers have 
\jaakko{used}
slight supervision \cite{Lee2010SemiSupervisedNM}, such as incorporating class label knowledge in semi-supervised approaches \cite{8820170} for 
\jaakko{downstream} tasks \cite{9013063}. Another study \cite{Haddock2020SemisupervisedNM} 
\jaakko{used}
maximum likelihood estimators under specific uncertainty distributions 
\jaakko{with}
multiplicative updates, showing flexibility across supervised tasks.



SeededLDA~\cite{Jagarlamudi2012} associates each topic with a seed set and biases topic assignment in documents containing matching seed words. KeyATM~\cite{Eshima} extends this by supporting topics without seeds and improving empirical robustness through selective seed specification and term weighting. Anchored CorEx~\cite{Gallagher2016AnchoredCE} takes an information-theoretic approach, anchoring seed words to specific topics. GuidedNMF~\cite{9413656} incorporates weighted seed supervision to guide topic formation and support tasks like classification~\cite{a15050136}. A recent semi-supervised variant~\cite{Lindstrom2022ContinuousSN} further combines prior knowledge with label information for improved latent topic discovery. However, such models often overfit to seed content, limiting generalization. \citet{10.1145/3539597.3570475} also proposed a seed-guided method based on contextual pattern alignment, but this can reinforce predefined structure and constrain topic diversity.

HGTM \cite{10.1007/978-3-031-66336-9_26} models rare topics 
\jaakko{with}
multiple topic-prevalence distributions, but 
\jaakko{needs}
topic-specific seed initialization. STM~\cite{pmlr-v28-das13} uses hierarchical Bayesian modeling with stick-breaking processes to capture low-frequency themes. Top2Vec \cite{Angelov2020Top2VecDR}, FASTopic \cite{wu2024fastopic}, and BERTopic \cite{grootendorst2022bertopic} have 
\jaakko{gained}
attention. However, they 
\jaakko{lack} ability to identify hidden topical patterns 
\jaakko{in}
a corpus \cite{Srivastava2017AutoencodingVI} when prevalences of topics or themes are imbalanced. 

Recent neural topic models improve interpretability by incorporating supervision or structural signals. SeededNTM~\cite{lin2023enhancing} uses multi-level seed word guidance at both word and document levels. NeuroMax~\cite{pham-etal-2024-neuromax} aligns topics with PLM-based embeddings through mutual information and optimal transport. Anchor-based models build unsupervised hierarchies by clustering seed anchors into interpretable trees~\cite{liu2024unsupervised}. Prior work has also explored covariate-based~\cite{eisenstein2011sparse, card-etal-2018-neural} and taxonomy-driven topic modeling~\cite{lee2022taxocom, lee2022taxonomyExpansion}, but such methods rely on structured metadata or external hierarchies. In contrast, our approach requires only a single seed list and mild constraints, enabling flexible recovery of minority topics from imbalanced data.

We focus on minority topics specific to our domain (e.g., mental health), rather than all rare content. This makes our task more challenging and sets our work apart from prior approaches~\cite{10.1007/978-3-031-66336-9_26, pmlr-v28-das13, 10.1145/3539597.3570475, Eshima, Haddock2020SemisupervisedNM}, etc.


\section{METHOD}
\label{sec:method}


Classical \jaakko{NMF} 
\cite{6165290, Egger2022ATM, Wu2024ASO}
creates a low-rank approximation of 
\jaakko{a}
non-negative valued 
matrix \( V \) representing a corpus. It is approximately decomposed as \( V \approx WH \) \cite{carbonetto2021non}.
The matrices \( W \) and \( H \) are fitted 
\jaakko{to minimize}
a reconstruction error objective function, here 
\jaakko{a}
generalized Kullback-Leibler (KL) divergence
\cite{Joyce2011KullbackLeiblerD}:

{\small
\begin{equation*}
D_{KL}(V \parallel WH) = \sum_{i,j} \Big( V_{ij} \log \frac{V_{ij}}
{(WH)_{ij}} 
- V_{ij} + (WH)_{ij} \Big)
\end{equation*}
}

Alternative forms of the penalty function are available, such as the Frobenius norm. Optimization iterates multiplicative update rules
similar to those by \citet{leelda2000algorithms, 4359171}. This process ensures \( W \) and \( H \) are adjusted to provide an optimal low-rank approximation of \( V \), while maintaining non-negativity of the factors. The classical
NMF update rules for \(W\) and \(H\) are:

{\small
\begin{equation*}
W_{ik} \leftarrow W_{ik} \frac{ \sum_{j'} \frac{H_{kj'} V_{ij'}}{(WH)_{ij'}} }{\sum_{j'} H_{kj'}} \;,\;
H_{kj} \leftarrow H_{kj} \frac{ \sum_{i'} \frac{ W_{i'k} V_{i'j}}{(WH)_{i'j}} }{\sum_{i'} W_{i'k}}
\end{equation*}
}


\subsection{Our Constrained NMF}
Let \( V \in \mathbb{R}^{M \times N} \) be
a document-term frequency matrix, where each row \( i \) corresponds to a document and each
column \( j \) represents the frequency of a word across all documents. Let \( W \in \mathbb{R}^{M \times K} \) be the document-topic distribution matrix, where \( K \) is the number of topics, and \( H \in \mathbb{R}^{K \times N} \) is the topic-word distribution matrix. In the context of topic modeling, \( W \) shows the distribution of topics across the documents in the corpus, and \( H \) captures the significance of terms across the topics.

Our model 
\jaakko{sets}
constraints on \( W \) and \( H \) to target minority themes while reducing noise. Constraints on \( W \) ensure documents without any seed words
do not 
\jaakko{have}
high prevalence of minority topics, while constraints on \( H \) align topics with domain-specific seed words, anchoring them to the content of interest. This dual technique allows effective minority topic detection without overfitting or emphasizing noise. Our new NMF method minimizes the generalized KL divergence with constraints on the low-rank matrices \( W \) and \( H \), given a user-defined seed word list. The constraints on \( W \) and \( H \) are 
\jaakko{detailed}
below; 
\jaakko{they} can be written as two sets of inequality constraints, \( g_1 \) and \( g_2 \), applied 
\jaakko{to} the two factor matrices. We show the constraints satisfy 
\jaakko{KKT} conditions \cite{Lange2013, ghojogh2021kkt}. 
\jaakko{Based on} the conditions,
we derive an optimization algorithm to find \( W \) and \( H \) minimizing the cost under the constraints. This yields a new multiplicative update rule 
\jaakko{differing}
from standard NMF. 

In our case study we use mental health discussion as a minority domain of interest, within YouTube comment data where the majority of content is not mental health related. We denote minority topics as 'mental health topics' for concreteness, but the method is general. The user can set the number \( K_{MH} \) of mental health topics (minority) to be modeled with guidance. The other $K-K_{MH}$ topics model other topical content: majority topics and minority content not interesting to the expert. $K_{MH}$ can be set as a desired level of detail in minority content but could also be chosen by typical model selection criteria. Guidance is provided as a set of seed words: a subset of terms known to be of interest in the minority domain \footnote{The seed word list provided in the GitHub repository consists of a single collection of mental health-related terms, without any predefined division into specific topics.}. We do not require the list to be comprehensive, and we do not require known divisions of words to predefined topics. We also do not require a known prior of minority content prevalences, or seed word prevalences. Thus the guidance is easy for experts to provide, and modeling will discover divisions and prevalences of minority topics in a data driven way. 

The constraints \( g_1 \) (Constraints on \( W \)) ensure that in documents where none of the known seed-words occur, prevalence of each minority topic 
\jaakko{should}
be at most an upper bound value. 
The constraints \( g_2 \) (Constraints on \( H \)) ensure that in each minority topic, at least some of the known-to-be-relevant seed words should have sufficient prevalence in the topic-to-word distribution, so that the total prevalence of the seed words \jaakko{is} 
at least a lower bound value. 
Both constraints are designed to make maximal use of the seed words, and separate minority topics from noise, while also allowing the model to discover how minority content is divided across topics, and without needing separate seed words for each topic to be discovered. 


\textbf{Constraint on $W$.}
Given the list of seed word indices $SI$, we identify the subset of documents not having any seed words, $I_0=\{i:\; \sum_{j \in SI} V_{ij} = 0\}$. 
For documents $i\in I_0$, we limit prevalence of each minority topic (in our case mental health topic) to at most a maximum \( W_{\text{max}} \) which can be set by the user or by model selection strategies.  
\jaakko{This yields}
per-element inequality constraints \( g_{1,ik} \),
%
%
%

 {\small
\begin{equation}
g_{1,ik}(W)  = W_{ik} - W_{\text{max}} \leq 0 
\;\;\; \forall k \in S_{MH},\; 
\forall i \in I_0
\label{eq:constraint_w}
\end{equation}
}

where 
for convenience 
\jaakko{we} denote minority topics as the first $K_{MH}$ topics $S_{MH}=\{1,\ldots,K_{MH}\}$. 
\jaakko{This} mild constraint 
states documents without seed words should not have \emph{strong} prevalence of minority topics; prevalence up to the user-set maximum is allowed. We do not 
\jaakko{set}
a converse constraint: documents having seed words are \emph{not} constrained to have high prevalence of minority topics.


\textbf{Constraint on $H$.}
We define mild constraints 
\( g_{2,k} \)
on seed word content in minority topics: 

{\small
\begin{equation}
g_{2,k}(H) = \theta_{\text{min}} - \frac{\sum_{j' \in SI} H_{kj'}}{\sum_{j'=1}^{N} H_{kj'}} \leq 0, \quad \forall k \in S_{MH} \;
\label{eq:constraint_h}
\end{equation}
}

The user can set \(\theta_{\text{min}} \) to control the focus on seed words in minority topics. We only use the mild overall constraint, and do not constrain prevalence of specific seed words per topic: which seed words may become prevalent in each minority topic is found by model fitting. We do not require a converse constraint, i.e., other topics are not required to avoid seed words.

\subsection{Lagrangian Formulation}
\texorpdfstring{The Lagrangian \cite{Leech1965}, \( L(W, H, \lambda, \mu) \)}{The Lagrangian (Leech 1965), L(W, H, lambda, mu)}, integrates the objective and the constraints. The objective is to minimize 
KL divergence \( D_{KL}(V \parallel WH) \) measuring 
how different the document-word matrix \( V \) is 
from the product of \( W \) (document-topic distribution) and \( H \) (topic-word distribution). 
Penalty terms are added to ensure constraints on \( W \) and \( H \) are met.
The Lagrangian \(L\) \jaakko{is:} 

{\small
\begin{equation}
L(W, H, \lambda, \mu) = D_{KL}(V \parallel WH) + \lambda \cdot g_1(W) + \mu \cdot g_2(H)
\label{eq:objectiveFunc}
\end{equation}
}

where \( g_1(W) \) and \( g_2(H) \) are sums over sets of constraints, and \(\lambda\) and \(\mu\) are Lagrange multipliers 
\jaakko{penalizing}
\jaakko{constraint violations.}
The \( g_1(W) \) sums constraints $g_{1,ik}(W)$ over indices \( i, k \), corresponding to elements of \( W \), while \( g_2(H) \) sums $g_{2,k}(H)$ over indices \( k \), corresponding to rows of \( H \).

\subsection{Karush-Kuhn-Tucker (KKT) Conditions}


\jaakko{An optimal solution must satisfy KKT conditions \cite{Lange2013, ghojogh2021kkt}  as follows.}

\textbf{Stationarity.}
To establish the stationarity condition in our optimization problem, we set gradients of the Lagrangian \( L \) to zero with respect to the \( W \) and \( H \). With respect to the KL divergence and inequality constraints \( g_1(W) \) and \( g_2(H) \), the requirements guarantee we are at a crucial point where changes in the cost are balanced. 

{\small
\begin{equation}
\frac{\partial L}{\partial W_{ik}} = \frac{\partial D_{KL}(V \parallel WH)}{\partial W_{ik}} + \lambda_{ik} \frac{\partial g_1(W)}{\partial W_{ik}} = 0 
\end{equation}
\begin{equation}
\frac{\partial L}{\partial H_{kj}} = \frac{\partial D_{KL}(V \parallel WH)}{\partial H_{kj}} + \mu_{k} \frac{\partial g_2(H)}{\partial H_{kj}} = 0
\end{equation}
}

\textbf{Primal Feasibility.}
For \( W \), this means the sum of specific elements (related to minority topics, like mental health) is capped by a maximum value \( W_{\text{max}} \). For \( H \), we ensure the proportion of seed words in each topic is
at least a user-defined minimum \( \theta_{\text{min}} \), as given in Eq. ~\eqref{eq:constraint_w}, and Eq. ~\eqref{eq:constraint_h}. This ensures the solution is within reasonable bounds and remains meaningful.



\textbf{Dual Feasibility.}
The Lagrange multipliers, \( \lambda \) and \( \mu \), must be non-negative. They denote strength of the penalty when a constraint is violated. 
If a constraint isn't violated, the multiplier can be zero. If the constraint is violated, the penalty pushes the solution back within the desired limits.

{\small
\begin{equation}
\lambda_{ik} \geq 0 \;\;,\;\;
\mu_{k} \geq 0
\end{equation}
}

\textbf{Complementary Slackness.}
If a constraint is already satisfied, there’s no need for a penalty. For instance, if elements in \( W \) for documents without seed words are below \( W_{\text{max}} \), the multiplier \( \lambda \) will be zero. Similarly, if the proportion of seed words in mental health topics exceeds \( \theta_{\text{min}} \), the multiplier \( \mu \) will be zero.
The product of each 
multiplier and its corresponding constraint must be zero:

{\small
\begin{equation}
\lambda_{ik} \left( \ W_{ik} - W_{\text{max}} \right) = 0\;, \;
\mu_{k}\left( \theta_{\text{min}} - \frac{\sum_{j' \in SI} H_{kj'}}{\sum_{j'=1}^{N} H_{kj'}} \right) = 0
\end{equation}
}

\noindent
To meet and solve the KKT criteria
we derive the gradients of above equations in Appendix \ref{sec:kl_divergence_gradient_wrt_W}- \ref{sec:multiplicative_update_for_H}.

\section{Optimization}
\label{sec:optimization}

We derive multiplicative update rules for the factorization matrices \( W \) and \( H \), to optimize the objective function under our constraints. 
Our Constrained NMF updates \( W \) and \( H \) iteratively.
Convergence properties of multiplicative update rules for unconstrained NMF have been studied \cite{gonzalez2005accelerating, BERRY2007155} including by \citet{4359171} for Euclidean distance and \citet{Finesso_2006} for generalized KL divergence. Convergence of multiplicative rules for NMF with inequality constraints remains open as the lifting approach of \citet{Finesso_2006} is not immediately applicable. 
Our updates derived using KKT conditions \cite{Lange2013,ghojogh2021kkt}
\jaakko{guarantee 
constraints are satisfied.}
In experiments our algorithm consistently demonstrated (Appendix \ref{sec:error_bars}, Figure~\ref{fig:error_bar}) to i) be nonincreasing with respect to the loss, and ii) converge to a stable point. 
Convergence theorems are left to future work.


We apply the KKT conditions \cite{Lange2013,ghojogh2021kkt} to handle the constraints such that optimal solutions \jaakko{satisfy}
non-negativity of the matrices and our 
\jaakko{domain} 
constraints. The multiplicative update rules
allow efficient updates
while retaining non-negativity of \( W \) and \( H \). The rules arise
from gradients of the Lagrangian, 
setting them to zero and solving for \( W \) and \( H \) iteratively. 
We derive update rules not only for elements $W_{ik}$ and $H_{kj}$ of \( W \) and \( H \) but also for Lagrange multipliers \( \lambda_{ik}\) and \( \mu_{k} \) that control satisfaction of the corresponding constraints (see Eqs. \ref{eq:lambW} \& \ref{eq:muH}).
To optimize the objective under our constraints, we update $W_{ik}$, $H_{kj}$, $\lambda_{ik}$, and $\mu_{k}$ 
\jaakko{by}
the rules.

\subsection{\texorpdfstring{The Multiplicative Update Rule for $W_{ik}$}{The multiplicative update rule for W\_ik}}

We set the Lagrangian gradient $\frac{\partial L}{\partial W_{ik}}$ to zero to satisfy the KKT condition \cite{Lange2013,ghojogh2021kkt}. Adding appendix Eq. \eqref{eq:klW} and Eq. \eqref{eq:g1_gradient_W}, we derive the gradient:

{\small
\begin{equation}
\frac{\partial L}{\partial W_{ik}}
= \sum_{j'} H_{kj'} \left( 1 - \frac{V_{ij'}}{(W H)_{ij'}} \right) 
+ \lambda_{ik}
\delta_{i \in I_0,\;k \in S_{MH}}
\end{equation}
}

\noindent
where the \(\delta\) function is 1 if \(k \in S_{MH}\) and \(i \in I_0\), and 0 otherwise.



\textbf{Final Multiplicative Update Rule for $W_{ik}$}:
To maintain non-negativity of $W_{ik}$, we solve $\frac{\partial L}{\partial W_{ik}}=0$ which yields that the ratio of the positive and negative terms in the gradient must be 1. This yields the multiplicative update for $W_{ik}$:

{\small
\begin{equation}
W_{ik} \leftarrow W_{ik}  \cdot \frac{\sum_{j'} H_{kj'} \frac{V_{ij'}}{(WH)_{ij'}}}{\sum_{j'} H_{kj'} +  \lambda_{ik} \delta_{i \in I_0,\;k \in S_{MH}}}
\label{eq:update_rule_for_W}
\end{equation}
}

\noindent
If the constraint \( g_{1,ik}(W) \) is active for some \( i\in I_0,\; k \in S_{MH} \), the Lagrange multiplier \( \lambda_{ik} \) adjusts the update to keep the constraint satisfied. If the constraint is not active, then \( \lambda_{ik} = 0 \).
The updating process is presented in Algorithm \ref{alg:mult_update_W}.

\subsection{\texorpdfstring{The Multiplicative Update Rule for $H_{kj}$}{The Multiplicative Update Rule for H\_kj}}

The update rule for $H_{kj}$ is derived from setting the gradient $\frac{\partial L}{\partial H_{kj}}$ of the Lagrangian to zero. By appendix Eq. \eqref{eq:klH} and Eq. \eqref{eq:g2H} the gradient becomes

{\small
\begin{multline}
\frac{\partial L}{\partial H_{kj}} = \sum_{i'} W_{i'k} \left( 1 - \frac{V_{i'j}}{(WH)_{i'j}} \right) \\
+ \mu_{k}  
\cdot
\delta_{k \in S_{MH}}\left(\frac{\text{Num}_k}{(\text{Den}_k)^2} -\delta_{j \in SI}\frac{1}{\text{Den}_k}\right)
\end{multline}
}

where $\text{Den}_k=\sum_{j'} H_{kj'}$, $\text{Num}_k=\sum_{j'\in SI} H_{kj'}$. 
Setting the above to zero, in the appendix we derive two multiplicative update rules, version 1 is provided below as
the \textbf{Final Multiplicative Update Rule for $H_{kj}$}:

{\small
\begin{equation}
    H_{kj} \leftarrow H_{kj} \cdot
    \frac{\sum_{i'} W_{i'k} \frac{V_{i'j}}{(WH)_{i'j}}}
    {\sum_{i'} W_{i'k} 
    + \mu_{k}  
    \cdot
    \delta_{k \in S_{MH}}\left(\frac{\text{Num}_k}{(\text{Den}_k)^2} -\frac{\delta_{j \in SI}}{\text{Den}_k}\right)}
    \label{eq:update_rule_for_H}
\end{equation}
}

See the updating process in Algorithm \ref{alg:mult_update_H}.




\subsection{Update Rules for Lagrange Multipliers}
Lagrange multipliers $\lambda$ and $\mu$ are updated based on whether the constraints are active or not. The $\lambda_{ik}$ and $\mu_{k}$ are updated using gradient ascent to ensure that they enforce the constraints on $W$ and $H$.

\textbf{Update Rule for \(\lambda_{ik}\).}\\
Given our constraint on $W$ given in Eq. \eqref{eq:constraint_w}:\\
If \(g_{1,ik}(W) < 0\): the constraint is inactive, we set \(\lambda_{ik} = 0\). \\
If \(g_{1,ik}(W) \ge 0\): the constraint is active, we update \(\lambda_{ik}\) using the following rule.

For active constraints, \(\lambda_{ik}\) is updated iteratively. One 
possible method 
a gradient ascent update:

{\small
\begin{equation}
\lambda_{ik}^{\text{new}} = \max\left(0, \lambda_{ik}^{\text{old}} + \eta \cdot g_{1,ik}(W)\right)
\label{eq:lambW}
\end{equation} 
}

where \(\eta\) is our learning rate, controlling how aggressively \(\lambda_{ik}\) is updated. The \(\max\) function ensures \(\lambda_{ik}\) stays non-negative, as required by our dual feasibility.

\textbf{Update Rule for \(\mu_{k}\).}\\
Recalling Eq. \eqref{eq:constraint_h}, given our constraint on $H$:\\
If \(g_{2,k}(H) < 0\):
The constraint is inactive, we set \(\mu_{k} = 0\). \\
If \(g_{2,k}(H) \ge 0\): 
The constraint is active, we update \(\mu_{k}\) using the following rule.

For active constraints, \(\mu_{k}\) is updated similarly to \(\lambda_{ik}\):
{\small
\begin{equation}
\mu_{k}^{\text{new}} = \max\left(0, \mu_{k}^{\text{old}} + \eta \cdot g_{2,k}(H)\right)
\label{eq:muH}
\end{equation}
}

During the \jaakko{optimization,}
the constraints are not 
\jaakko{satisfied} at every iteration. 
\jaakko{Each} Lagrange multiplier ($\lambda_{ik}$ and $\mu_k$) becomes active (nonzero) 
\jaakko{if}
its corresponding constraint is violated. 
\jaakko{The algorithm thus}
adjusts $W$ and $H$ iteratively, striving to balance 
satisfaction of the constraints with minimizing the primary objective function. 
Eq. \eqref{eq:update_rule_for_W} to Eq. \eqref{eq:muH} provide easy-to-implement updating rules. Until the objective value of Eq. \eqref{eq:objectiveFunc} remains unchanged, we iteratively modify \(W\) and \(H\). Algorithm \ref{alg:optimization_WH} gives an outline of this approach. The time complexity of our model is detailed in the Appendix \ref{sec:hyperparameter_tuning_details}.
We demonstrate the strength of our method in the \jaakko{next}
section by comparing its performance on synthetic data to SOTA models \jaakko{and} carrying out a case study.
\begin{algorithm}[ht]
\footnotesize 
\caption{Multiplicative Update Rule for the \(W_{ik}\)}
\label{alg:mult_update_W}
\resizebox{\columnwidth}{!}{%
\begin{minipage}{\columnwidth}
\begin{algorithmic}
   \STATE {\bfseries Input:} {\small $V \in \mathbb{R}^{M \times N}$, $W \in \mathbb{R}^{M \times K}$, $H \in \mathbb{R}^{K \times N}$, $\lambda_{ik}$ (Lagrange multipliers for $g_{1,ik}(W)$), $W_{\text{max}}$ (upper bound constraint)}
   \STATE {\bfseries Output:} {\small Updated $W_{ik}$}
   \REPEAT
   \FOR{each $i \in [1, M]$ and $k \in [1, K]$}
      \STATE numerator $\leftarrow \sum_{j'} H_{kj'} \cdot \left( \frac{V_{ij'}}{(WH)_{ij'}} \right)$
      \STATE denominator $\leftarrow \sum_{j'} H_{kj'}$
      \IF{$i \in I_0$ and $k \in S_{MH}$}
         \STATE Add $\lambda_{ik}$ to denominator
      \ENDIF
      \STATE $W_{ik} \leftarrow W_{ik} \cdot \frac{\text{numerator}}{\text{denominator}}$
   \ENDFOR
   \UNTIL{convergence}
\end{algorithmic}
\end{minipage}
}
\end{algorithm}

\begin{algorithm}[ht]
\footnotesize 
\caption{Multiplicative Update Rule for the \(H_{kj}\)}
\label{alg:mult_update_H}
\resizebox{\columnwidth}{!}{%
\begin{minipage}{\columnwidth}
\begin{algorithmic}
   \STATE {\bfseries Input:} {\small $V \in \mathbb{R}^{M \times N}$, $W \in \mathbb{R}^{M \times K}$, $H \in \mathbb{R}^{K \times N}$, $\mu_{k}$, seed indices, \(S_{MH}\), $\theta_{\text{min}}$}
   \STATE {\bfseries Output:} {\small Updated $H_{kj}$}
   \FOR{each $k \in [1, K]$ and $j \in [1, N]$}
      \STATE numerator $\leftarrow \sum_{i'} W_{i'k} \cdot \left( \frac{V_{i'j}}{(WH)_{i'j}} \right)$
      \STATE denominator $\leftarrow \sum_{i'} W_{i'k}$
      \IF{$k \in S_{MH}$}
         \STATE Compute constraint term:
         \STATE constraint $\leftarrow \mu_k \cdot \left( \frac{\text{Num}_k}{(\text{Den}_k)^2} - \delta_{j \in SI} \cdot \frac{1}{\text{Den}_k} \right)$
         \STATE denominator $\leftarrow$ denominator $+$ constraint
      \ENDIF
      \STATE $H_{kj} \leftarrow H_{kj} \cdot \frac{\text{numerator}}{\text{denominator}}$
   \ENDFOR
\end{algorithmic}
\end{minipage}
}
\end{algorithm}
\begin{algorithm}[ht]
\footnotesize 
\caption{Optimization Procedure for \(W\) and \(H\)}
\label{alg:optimization_WH}
\resizebox{\columnwidth}{!}{%
\begin{minipage}{\columnwidth}
\begin{algorithmic}
   \STATE {\bfseries Input:} $V$, initial $W$, $H$, constraints $g_1(W)$, $g_2(H)$
   \STATE {\bfseries Output:} Optimized $W$, $H$
   \STATE Initialize $\lambda_{ik}$, $\mu_k$ (Lagrange multipliers)
   \REPEAT
      \STATE Update $W_{ik}$ via Alg.~\ref{alg:mult_update_W}
      \STATE Update $H_{kj}$ via Alg.~\ref{alg:mult_update_H}
      \STATE Check $g_1(W) \leq 0$ and $g_2(H) \leq 0$
      \STATE \textbf{Update} $\lambda_{ik}$ (see Eq.~\ref{eq:constraint_w}):
      \IF{$g_{1,ik}(W) < 0$}
         \STATE $\lambda_{ik} \leftarrow 0$ \COMMENT{Inactive constraint}
      \ELSE
         \STATE Update $\lambda_{ik}$ via Eq.~\ref{eq:lambW}
      \ENDIF
      \STATE \textbf{Update} $\mu_k$ (see Eq.~\ref{eq:constraint_h}):
      \IF{$g_{2,k}(H) < 0$}
         \STATE $\mu_k \leftarrow 0$ \COMMENT{Inactive constraint}
      \ELSE
         \STATE Update $\mu_k$ via Eq.~\ref{eq:muH}
      \ENDIF
   \UNTIL{convergence or stopping condition is met}
   \STATE \textbf{Return :} optimized $W$, $H$
\end{algorithmic}
\end{minipage}
}
\end{algorithm}



\section{Experiments}
\label{sec:experiments}

We evaluate our model on synthetic and real-world Finnish YouTube data. 
We benchmark against a wide range of conventional and neural topic models, including NMF~\cite{leelda2000algorithms}, LDA~\cite{10.5555/944919.944937}, ProdLDA~\cite{Srivastava2017AutoencodingVI}, Top2Vec~\cite{Angelov2020Top2VecDR}, BERTopic~\cite{grootendorst2022bertopic}, FASTopic~\cite{wu2024fastopic}, KeyATM~\cite{Eshima}, GuidedNMF~\cite{9413656}, GuidedLDA and SeededLDA~\cite{Jagarlamudi2012}, and Corex~\cite{Gallagher2016AnchoredCE}. We used publicly available implementations (see Appendix~\ref{sec:hyperparameter_tuning_details}), with shared preprocessing, seed word list, and a uniform topic count \footnote{We used default hyperparameters unless otherwise noted, tuning only when required for baseline stability or fairness.}. 

\subsection{Datasets}
\label{sec:dataset}
\textbf{Real Dataset.}
We selected~20 Finnish YouTubers recommended by a public health expert, focusing on mental health content for younger audiences. We scraped comments and metadata from their videos up to March 15, 2024. This yielded roughly~5.5 million Finnish-language comments. While most comments cover various topics, mental health discussions are a minority. Table~\ref{tab:dataset_stats} summarizes statistics of both datasets.

\textbf{Synthetic Dataset.}
To build a synthetic dataset, topic-related words were injected into randomly selected sentences to simulate mental health discussions. A synthetic ground truth of 18 mental health topics was defined, each with related Finnish words. Topics like \textit{Suicide} and \textit{Anxiety} had terms such as \textit{“itsemurha”} (suicide) and \textit{“ahdistus”} (anxiety), while others like \textit{Mental Health} and \textit{Social Isolation} included words like \textit{“yksinäisyys”} (loneliness) and \textit{“trauma.”} 
%
%
To generate the data, we randomly sampled 500 documents. Each had a 10\% chance 
\jaakko{to receive}
mental health content: if selected, one synthetic topic was randomly assigned, and four related words were injected. The remaining documents were left unchanged and labeled “-1” to indicate no mental health topic. This created a realistic mix of general and topic-specific content.


\begin{table}[t]
    \centering
    \footnotesize
    \begin{tabular}{lll}
    \hline
    \textbf{Statistic} & \textbf{Real} & \textbf{Synthetic} \\
    \hline
    Documents (raw) & 5,578,289 & 500 \\
    Documents (filtered) & 2,979,969 & 500 \\
    Avg. words/doc & 7.04 & 10 \\
    Vocabulary size & 885,945 & 2,800 \\
    \hline
    \end{tabular}
    \caption{Dataset statistics after preprocessing.}
    \label{tab:dataset_stats}
\end{table}


\subsection{Evaluation Metrics}
\label{sec:evaluation}


We evaluate clustering quality 
by
Purity and normalized mutual information (NMI) ~\cite{christopher2008introduction} following \cite{wu2023effective, zhao2021neural}. 
We further
introduce a customized purity function that 
focuses on
true minority topics ignoring background (non-mental-health) labels. Standard Purity reflects majority content, which is not our goal. To address this, we compute a focused purity score 
measuring only
accuracy of minority topics.
In the synthetic dataset, each document \(d = 1, \ldots, M\) is either injected with minority-related (mental health) words from one ground-truth minority topic (labeled \(y_d = 1, \ldots, Y\)) or (labeled \(y_d = -1\)) for background. The model assigns each document \(d\) to its most probable predicted topic \(t_d\). For each predicted topic $k$, we count how many documents are assigned; 
$\text{count}(k) = \sum_{d=1}^{M} \delta(t_d = k)$, where $\delta(\cdot)$ is the indicator function. Similarly, how many of those belong to a ground-truth minority label $y$; $\text{count}(y, k) = \sum_{d=1}^{M} \delta(y_d = y, \, t_d = k)$. 
The dominant valid label for each topic $k$ is defined as \(y^k = \arg\max_{y > -1} \text{count}(y, k)\), excluding the background. The purity score is computed as:

{\small
\[
\text{Purity} = \frac{\sum_{k=1}^{K} \text{count}(y^k, k)}{\sum_{k=1}^{K} \delta(\text{count}(y^k, k) > 0) \cdot \text{count}(k)}
\label{eq:purity_score}
\]
}


To assess topic quality beyond clustering, we compute JSD between injected ground-truth topic-word distributions and the model’s learned topics. For each ground-truth topic, we report the minimum JSD to its closest learned topic:
\[
\min_k \text{JSD}(P_{\text{true}}, P^{(k)}_{\text{model}})
\]
JSD 
is 
more reliable
for evaluating low-prevalence content than standard coherence metrics\jaakko{, as it compares to ground-truth minority content.}

\subsection{Results}
\label{sec:results}
On the synthetic set the models were 
\jaakko{tasked}
to find 20 topics (7 minority and 13 majority).
Conventional NMF, LDA, Top2Vec, FASTopic, ProdLDA were run without seed word guidance. Corex, GuidedLDA, SeededLDA, BERTopic, KeyATM, and GuidedNMF were run with the same seed words as our model's setting. 
The quality of the models was then evaluated by the metrics discussed above.

\begin{figure}[ht]
    \footnotesize
    \setlength{\tabcolsep}{1.5pt}
\begin{minipage}[c]{3.6cm}
    \begin{tabular}{lcc}
    \hline
    \textbf{Model} & \textbf{Purity} & \textbf{NMI} \\
    \hline
    NMF         & 0.0969 & 0.1846 \\
    LDA         & 0.0500 & 0.1131 \\
    Corex       & 0.0677 & 0.1697 \\
    GuidedLDA   & 0.0635 & 0.1721 \\
    SeededLDA   & 0.0923 & 0.1660 \\
    KeyATM      & 0.0878 & 0.1791 \\
    GuidedNMF   & 0.0723 & 0.2247 \\
    Top2Vec     & 0.0184 & 0.1055 \\
    FASTopic    & 0.0762 & 0.1617 \\
    BERTopic    & 0.0381 & 0.0914 \\
    ProdLDA     & 0.0577 & 0.1104 \\
    \textbf{Ours} & \textbf{0.1765} & \textbf{0.2453} \\
    \hline
    \end{tabular}
\end{minipage}
\begin{minipage}[c]{4.0cm}
\includegraphics[width=\columnwidth]{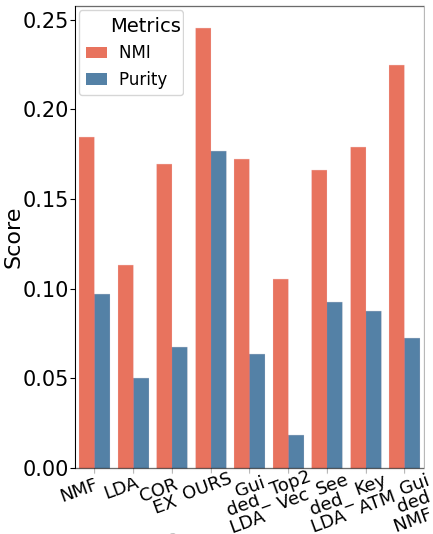}
\end{minipage}
\caption{Comparison of NMI and Purity Scores across Baselines on synthetic dataset (20 topics, 7 mental health topics, 500 samples). Left: Result table, the best is in \textbf{bold}.
Right: results as a bar graph.}

\label{fig:Purity_and_NMI_table_and_figure}
\end{figure}

Figure \ref{fig:Purity_and_NMI_table_and_figure}
reports NMI and Purity scores 
across 
the
models on the synthetic dataset.
\footnote{We do not report NMI, Purity, or JSD for the real dataset, as it lacks ground-truth topic labels, but we present the discovered topics for interpretability.} 
Our model achieves the highest scores on both metrics and outperforms all the baselines, including both unsupervised and seed-guided approaches. Furthermore, our approach guided the model to discover latent themes within the topics, rather than forcing seed words into the content, unlike models such as Corex, GuidedLDA, and SeededLDA. Notably, an inspection of the results revealed that their strategy predominantly incorporated seed words directly into the discovered topics, resulting in poor interpretability and performance. 
We also performed additional analyses on using various numbers of topics (see Appendix~\ref{appendix:topic-count-analysis}, Figures \ref{fig:nmi_scores(30,50,80)}, \ref{fig:purity_scores(30,50,80)}, and \ref{fig:purity_nmi_scatter}, again showing good performance of our model.)

\begin{figure}[ht]
\centering
\includegraphics[width=\linewidth]{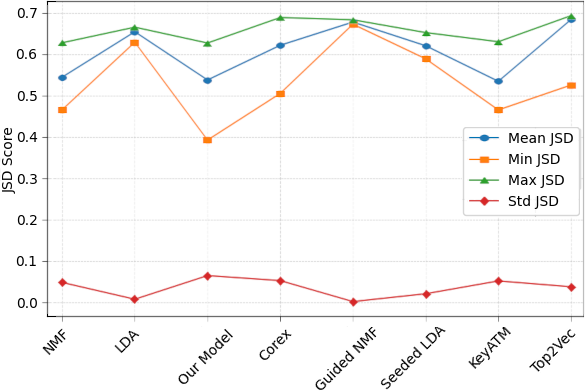}
\caption{Topic Quality using JSD Score}
\label{fig:jsd_evaluation_small}
\end{figure}

Figure~\ref{fig:jsd_evaluation_small} reports the topic quality of the models according to JSD. 
Here as well, our method yielded best performance 
(smallest JSDs) in discovering minority themes. Thus we achieve both high topic quality (small JSD) and high clustering ability (high purity \& NMI) outperforming others.

The detailed topics discovered by our model are shown in Appendix~\ref{sec:discovered_topics}, Table~\ref{tab:tableA6}.



\subsection{Ablation Study}
We 
\jaakko{ran}
an ablation study on the synthetic dataset to 
assess
impact of modeling choices. 
We varied the number of total topics (30, 50, 80), number of minority topics (10, 15, 20), and hyperparameters \(W_{\text{max}}\) and \(\theta_{\text{min}}\). We assessed their effect on KL divergence, NMI, and purity. Our model 
is
robust and outperformed baselines 
on
all settings. Detailed results are 
in Appendix~\ref{appendix:sensitivity} in Figures \ref{fig:kl_sensitivity}, \ref{fig:nmi_purity_sensitivity}.

\section{Discovered Topics}
\label{sec:findings}


We present example outputs from our Constrained NMF.
Topics discovered in the real data include highly meaningful mental health concerns such as \emph{How mental health differs from outward appearance} (Topic 0, top words crazy, appearance, medicine), \emph{How mental health problems may exacerbate around holidays} (Topic 1; expectation, christmas, mental health problem), \emph{Sadness and suicide} (Topic 2; sad, suicide, human), and \emph{Support and ADHD} (Topic 3; to support, adhd, crisis).
Full lists in Finnish with English translations for clarity are provided in the Appendix: discovered topics on the synthetic dataset in Appendix~\ref{sec:discovered_topics}, Table~\ref{tab:tableA6}, and Tables~\ref{tab:tableA2}--\ref{tab:tableA5} for our real-world YouTube dataset.


\section{Conclusions and Future Work}
\label{sec:conclusions}

We introduced a constrained NMF model for discovering domain-specific minority topics without 
\jaakko{explicit} topic-level supervision. 
\jaakko{We use}
soft prevalence constraints and a single seed word list to guide discovery of distinct, data-driven minority themes. In experiments on synthetic and real-world YouTube comment data, our method outperforms strong baselines in clustering and topic quality, 
successfully modeling low-prevalence mental health discussions.
This 
shows
potential of constrained matrix factorization to identify 
patterns in noisy, imbalanced corpora. Future work includes expanding to new domains and exploring integration with neural or contextualized topic models.
\section{Limitations}
\label{sec:limitations}
Evaluating low-frequency topics remains challenging, as standard metrics often fail to reflect minority theme quality. While we use synthetic ground truth, real-world datasets lack annotated minority themes, and benchmarks in domains like mental health are scarce. To estimate the quality of a topic, we compute the JSD in discovered and ground-truth topic distributions in synthetic settings. Although, NMI, purity, JSD and similar automated measures are informative, these methods risk being misaligned or biased when it comes to the value of meaning~\cite{hoyle2021automated}. More reliable measures of semantic value include human assessment, which can enhance evaluation of topic coherence and relevance in so-called minority content. Future work should address this gap.


\bibliographystyle{acl_natbib}
\bibliography{custom}

\appendix
\section{Appendix}
\label{sec:the_appendix}

This Appendix contains several derivations and details complementing the main paper, as follows.

We provide derivatives (gradients) of our generalized Kullback-Leibler divergence loss function in Sections \ref{sec:kl_divergence_gradient_wrt_W} and \ref{sec:kl_divergence_gradient_wrt_H}, derivatives of the constraint functions in Sections
\ref{sec:g1_gradient_wrt_W} and \ref{sec:g2_gradient_wrt_H}, 
and derivatives of the full Lagrangian in Sections \ref{sec:lagrangian_gradient_wrt_W} and \ref{sec:lagrangian_gradient_wrt_H}. 
The derivatives are used to solve multiplicative update rules satisfying the Karush-Kuhn-Tucker conditions; we provide the details in Sections
\ref{sec:multiplicative_update_for_W} and
\ref{sec:multiplicative_update_for_H}.
Next, Sections \ref{sec:details_of_the_datasets}, \ref{sec:hyperparameter_tuning_details}, and \ref{sec:training_details} include additional details, such as details concerning the training datasets, hyperparameter choices, the computer system utilized, time complexity, and training details.
Section~\ref{sec:discovered_topics} provides additional detailed experimental results for the synthetic and real data experiments, showing topics discovered by our proposed method in terms of their top words. Section~\ref{sec:error_bars} provides error bars from multiple runs. Lastly, Appendix~\ref{appendix:sensitivity} provides an ablation study on the synthetic dataset, evaluating varying hyperparameter settings on different metrics, and Appendix~\ref{appendix:topic-count-analysis} provides additional analysis on how varying the number of topics counts, minority supervision impacts NMI and purity, along with an evaluation of topic quality using JSD across several baselines.   



\subsection{\texorpdfstring{KL Divergence Gradient with Respect to $W_{ik}$}{KL Divergence Gradient with Respect to W\_ik}}
\label{sec:kl_divergence_gradient_wrt_W}


As stated in Section \ref{sec:method} of the main paper, the generalized Kullback-Leibler (KL) divergence between $V$ and $WH$ is


{\small
\begin{align*}
D_{KL}(V \parallel WH) = \sum_{i',j'} \bigg(
& V_{i'j'} \log \frac{V_{i'j'}}{(WH)_{i'j'}} \\
& - V_{i'j'} + (WH)_{i'j'} \bigg)
\end{align*}
}

where we use $i'$ and $j'$ as sum indices for clarity, to distinguish them from indices of specific elements whose derivatives we will compute.

The generalized KL divergence is a special case of a Bregman divergence. While the classical KL divergence compares two probability distributions whose elements are probabilities that sum to 1, the generalized KL divergence applies between two sets of nonnegative numbers whose elements do not need to sum to 1: here the sets are the elements of $V$ and the corresponding elements of $WH$.


To find the derivative $\frac{\partial D_{KL}(V \parallel WH)}{\partial W_{ik}}$ with respect to an element $W_{ik}$, we will differentiate each term in $D_{KL}$. 

First, recall:

{\small
\begin{equation*}
(WH)_{i'j'} = \sum_{k'} W_{i'k'} H_{k'j'} \;.
\end{equation*}
}

The derivative of $(WH)_{i'j'}$ with respect to $W_{ik}$ is then

{\small
\begin{equation*}
\frac{\partial}{\partial W_{ik}}  (W H)_{i'j'}  = \frac{\partial}{\partial W_{ik}}  \sum_{k'} W_{i'k'} H_{k'j'} = \delta_{i'i} H_{kj'}
\end{equation*}
}

where \(\delta_{i'i} = 1\) if \(i'=i\) and zero otherwise.

The first term inside the sum in the KL divergence is $V_{i'j'} \log \frac{V_{i'j'}}{(WH)_{i'j'}}$. Its derivative with respect to $W_{ik}$ is


{\small
\begin{multline*}
\frac{\partial}{\partial W_{ik}} \left( V_{i'j'} \log \left( \frac{V_{i'j'}}{(WH)_{i'j'}} \right) \right)
= \\ - \left( \frac{V_{i'j'}}{(WH)_{i'j'}} \right) 
\left( \frac{\partial}{\partial W_{ik}} (WH)_{i'j'} \right)
= \\ - \left( \frac{V_{i'j'}}{(WH)_{i'j'}} \right)
\left( \delta_{i'i} H_{kj'} \right)
\end{multline*}
}

Summing this over $i'$ and $j'$, we get the derivative of the first part of the sum:

\noindent
{\small
\begin{multline*}
\frac{\partial}{\partial W_{ik}} \sum_{i'j'} \left(V_{i'j'} \log \left[ \frac{V_{i'j'}}{(W H)_{i'j'}} \right] \right) 
= \\ \sum_{i'j'}  -\left( \frac{V_{i'j'}}{(W H)_{i'j'}} \right) \left( \delta_{i'i} H_{kj'} \right) 
= \\ \sum_{j'}  -\left( \frac{V_{ij'}}{(W H)_{ij'}} \right)  H_{kj'} 
= - \sum_{j'} \frac{V_{ij'} H_{kj'}}{(W H)_{ij'}}
\end{multline*}
}

The second term inside the sum in the KL divergence is $-V_{i'j'}$ which is constant with respect to $W_{ik}$ hence its derivative is zero.

The third term is $(WH)_{i'j'}$ itself. Its derivative with respect to $W_{ik}$ was already derived above to be

{\small
\begin{equation*}
\frac{\partial (WH)_{i'j'}}{\partial W_{ik}} = \delta_{i'i} H_{kj'} \;.
\end{equation*}
}

Summing this over $i'$ and $j'$, we get the derivative of the second part of the sum:


{\small
\begin{multline*}
\frac{\partial}{\partial W_{ik}}  \sum_{i'j'} (W H)_{i'j'}
= \\ \sum_{i'j'}  \frac{\partial}{\partial W_{ik}} (W H)_{i'j'}
= \sum_{i'j'}  \delta_{i'i} H_{kj'}
= \sum_{j'} H_{kj'} \;.
\end{multline*}
}

\textbf{Combining the Derivatives:} 
The sum of the two parts above yields the full derivative of $D_{KL}(V \parallel WH)$ with respect to $W_{ik}$ as Eq.~\eqref{eq:klW}:


{\small
\begin{multline}
\frac{\partial}{\partial W_{ik}} D_{KL}(V \parallel W H)
= \sum_{j'} H_{kj'} -  \sum_{j'}  \frac{V_{ij'} H_{kj'}}{(W H)_{ij'}}
= \\ \sum_{j'} H_{kj'} \left( 1 - \frac{V_{ij'}}{(W H)_{ij'}} \right) \;.
\label{eq:klW}
\end{multline}
}

\subsection{\texorpdfstring{KL Divergence Gradient with Respect to $H_{kj}$}{KL Divergence Gradient with Respect to H\_kj}}
\label{sec:kl_divergence_gradient_wrt_H}

Similarly, we differentiate the KL divergence with respect to $H_{kj}$. Firstly, the derivative of $(WH)_{i'j'}$ with respect to $H_{kj}$ is

{\small
\begin{equation*}
\frac{\partial}{\partial H_{kj}}  (W H)_{i'j'}  = \frac{\partial}{\partial H_{kj}}  \sum_{k'} W_{i'k'} H_{k'j'} = \delta_{j'j} W_{i'k}
\end{equation*}
}

where \(\delta_{j'j} = 1\) if \(j'=j\) and zero otherwise. The first term inside the sum of the $D_{KL}$ is $V_{i'j'} \log \frac{V_{i'j'}}{(WH)_{i'j'}}$
and its derivative becomes


{\small
\begin{multline*}
\frac{\partial}{\partial H_{kj}}  \left(V_{i'j'} \log \left( \frac{V_{i'j'}}{(W H)_{i'j'}} \right) \right)
=  \\ - \frac{V_{i'j'}}{(W H)_{i'j'}} \left( \frac{\partial}{\partial H_{kj}} (W H)_{i'j'} \right)
=  \\ - \frac{V_{i'j'}}{(W H)_{i'j'}} \left( \delta_{j'j} W_{i'k} \right)
\end{multline*}
}

Summing over $i$ and $j$, we get the derivative of the first part of the $D_{KL}$ sum:


{\small
\begin{multline*}
\frac{\partial}{\partial H_{kj}}  \sum_{i'j'} \left(V_{i'j'} \log \left( \frac{V_{i'j'}}{(W H)_{i'j'}} \right)\right)
= \\ \sum_{i'j'}  \left(-\frac{V_{i'j'}}{(W H)_{i'j'}} \right) \delta_{j'j} W_{i'k} 
= - \sum_{i'}  \frac{V_{i'j} W_{i'k}}{(W H)_{i'j}} \;.
\end{multline*}
}

The second term inside the sum in the KL divergence is $-V_{i'j'}$ is constant with respect to $H_{kj}$ hence its derivative is zero.

The derivative of the third part of the $D_{KL}$ sum becomes

{\small
\begin{equation*}
\frac{\partial}{\partial H_{kj}}  \sum_{i'j'} (W H)_{i'j'}
= \sum_{i'j'}   \left( \delta_{j'j} W_{i'k} \right)
= \sum_{i'}  W_{i'k} \;.
\end{equation*}
}

\textbf{Combining the derivatives:}
Adding the parts together, the derivative of $D_{KL}$ with respect to $H_{kj}$ is


{\small
\begin{multline}
\frac{\partial}{\partial H_{kj}} D_{KL}(V \parallel W H) 
= \\ \sum_{i'}  W_{i'k}  - \sum_{i'}   \frac{V_{i'j} W_{i'k}}{(W H)_{i'j}} 
= \\ \sum_{i'}  W_{i'k} \left( 1 - \frac{V_{i'j}}{(W H)_{i'j}} \right) \;.
\label{eq:klH}
\end{multline}
}

\subsection{\texorpdfstring{Gradient of the constraints $g_1(W)$}{Gradient of the constraints g\_1(W)}}
\label{sec:g1_gradient_wrt_W}

The constraints $g_1(W)$ are given for each element of $W$ by

{\small
\begin{equation*}
g_{1,ik}(W) = W_{ik} - W_{\text{max}} \leq 0 \quad \forall i \in I_0, \forall k \in S_{MH} \;.
\end{equation*}
}

The constraints \( g_1(W) \) ensure that each weight \( W_{ik} \) individually does not exceed the maximum value \( W_{\text{max}} \). 


\textbf{Derivative of the constraints $g_1(W)$ with respect to $W_{ik}$.}
It is easy to see each individual constraint affects only one element $W_{ik}$ of $W$, hence the constraint has a nonzero derivative only with respect to that element. Thus the sum of the constraints $g_1(W)$ has a nonzero derivative only if $W_{ik}$ is one of the constrained elements.
The derivative is


{\small
\begin{multline}
\frac{\partial g_1(W)}{\partial W_{ik}} 
= \frac{\partial}{\partial W_{ik}} \sum_{i' \in I_0, k'\in S_{MH}} (W_{i'k'} - W_{\text{max}}) 
= \\ \begin{cases} 
1 & \text{if } i \in I_0, k \in S_{MH} \\
0 & \text{otherwise}
\end{cases}
\label{eq:g1_gradient_W}
\end{multline}
}

where \( S_{MH} \) is the subset of mental health topics, and the derivative indicates that the constraint is active when \( i \in I_0 \) and \( k \in S_{MH} \), and inactive otherwise.

\subsection{\texorpdfstring{Gradient of the constraints $g_2(H)$}{Gradient of the constraints g\_2(H)}}
\label{sec:g2_gradient_wrt_H}

The constraints $g_2(H)$ are given for each element of $H$ by

{\small
\begin{equation*}
g_{2,k}(H) = \theta_{\text{min}} - \frac{\sum_{j' \in SI} H_{kj'}}{\sum_{j'=1}^{N} H_{kj'}} \leq 0,\; \forall k \in S_{MH}
\end{equation*}
}

\textbf{Derivative of the constraints $g_2(H)$ with respect to $H_{kj}$.}
It is easy to see that each individual constraint $g_{2,k}(H)$ concerns all elements $H_{kj}$ in a particular row $k$ of $H$ hence the constraint has a nonzero derivative only with respect to elements in that row.

We first define helper notations \(\text{Num}\) and \(\text{Denom}\) which are also used in the main paper to state the Final Multiplicative Update Rule for \(H_{kj}\). The detailed definitions were left in the appendix due to space limitations, and we provide them here.


\textbf{Step 1: We define the Components first}

Let $\text{Num}_{k}$ be the numerator:

{\small
\begin{equation*}
\text{Num}_{k} = \sum_{j' \in SI} H_{kj'}
\end{equation*}
}

Let $\text{Den}_{k}$ be the denominator:

{\small
\begin{equation*}
\text{Den}_{k} = \sum_{j'=1}^{N} H_{kj'}
\end{equation*}
}

Thus, we can rewrite the constraint as:

{\small
\begin{equation*}
g_{2,k}(H) = \theta_{\text{min}} - \frac{\text{Num}_k}{\text{Den}_k}
\end{equation*}
}

\textbf{Step 2: The Quotient Rule}

To find the derivative of $g_{2,k}(H)$ with respect to $H_{kj}$, we use the quotient rule:

{\small
\begin{equation*}
\frac{\partial g_{2,k}(H)}{\partial H_{kj}} = -\frac{\frac{\partial (\text{Num}_k)}{\partial H_{kj}} \cdot \text{Den}_k - \text{Num}_k \cdot \frac{\partial (\text{Den}_k)}{\partial H_{kj}}}{(\text{Den}_k)^2}
\end{equation*}
}

\textit{Partial derivative of $\text{Num}_k$ with respect to $H_{kj}$}:

{\small
\begin{align*}
\frac{\partial (\text{Num}_k)}{\partial H_{kj}} = 
\begin{cases}
1 & \text{if } k \in S_{MH}, j \in SI\\
0 & \text{otherwise}
\end{cases}
\end{align*}
}

\textit{Partial derivative of $\text{Den}_k$ with respect to $H_{kj}$}:

{\small
\begin{align*}
\frac{\partial (\text{Den}_k)}{\partial H_{kj}} 
=\begin{cases}
1 & \text{if } k \in S_{MH}\\
0 & \text{otherwise}
\end{cases}
\end{align*}
}

\textbf{Step 3: Substituting Back into the Quotient Rule}

If $j \in SI$, $k \in S_{MH}$:

{\small
\begin{equation*}
\frac{\partial g_{2,k}(H)}{\partial H_{kj}} = -\frac{1 \cdot \text{Den}_k - \text{Num}_k \cdot 1}{(\text{Den}_k)^2} = -\frac{\text{Den}_k - \text{Num}_k}{(\text{Den}_k)^2}
\end{equation*}
}

If $j \notin SI$, $k \in S_{MH}$:

{\small
\begin{equation*}
\frac{\partial g_{2,k}(H)}{\partial H_{kl}} = -\frac{0 \cdot \text{Den}_k - \text{Num}_k \cdot 1}{(\text{Den}_k)^2} = \frac{\text{Num}_k}{(\text{Den}_k)^2}
\end{equation*}
}

Thus, the final derivative is:

{\small
\begin{equation}
\frac{\partial g_{2,k}(H)}{\partial H_{kj}} =
\begin{cases}
-\frac{\text{Den}_k - \text{Num}_k}{(\text{Den}_k)^2} & \text{if } k \in S_{MH},\; j \in SI\\
\frac{\text{Num}_k}{(\text{Den}_k)^2} & \text{if } k \in S_{MH},\;j \notin SI\\
0 & \text{otherwise}
\end{cases}
\label{eq:g2H}
\end{equation}
}

\subsection{\texorpdfstring{Lagrangian Derivative with Respect to $W_{ik}$}{Lagrangian Derivative with Respect to W\_ik}}
\label{sec:lagrangian_gradient_wrt_W}

The Lagrangian is:


{\small
\begin{multline*}
L(W, H, \lambda, \mu) = D_{KL}(V \parallel WH) + \\ \sum_{i',k'} \lambda_{i'k'} g_{1,i'k'}(W) + \sum_{k'} \mu_{k'} g_{2,k'}(H)
\end{multline*}
}

Taking the derivative with respect to $W_{ik}$, and noting that the only constraint affected by $W_{ik}$ is $g_{1,ik}(W)$:

{\small
\begin{equation*}
\frac{\partial L}{\partial W_{ik}} = \frac{\partial D_{KL}(V \parallel WH)}{\partial W_{ik}} + \lambda_{ik} \frac{\partial g_{1,ik}(W)}{\partial W_{ik}} 
\end{equation*}
}
\\
Inserting the results from Eqs.~ \eqref{eq:klW} and \eqref{eq:g1_gradient_W}, we get:


{\small
\begin{multline}
\frac{\partial L}{\partial W_{ik}}
= \sum_{j'} H_{kj'} \left( 1 - \frac{V_{ij'}}{(W H)_{ij'}} \right) 
+ \\ \lambda_{ik} \times
\begin{cases}
1 & \text{if } i \in I_0, k \in S_{MH} \\
0 & \text{otherwise}
\end{cases}
\end{multline}
}

\subsection{\texorpdfstring{Lagrangian Derivative with Respect to $H_{kj}$}{Lagrangian Derivative with Respect to H\_kj}}
\label{sec:lagrangian_gradient_wrt_H}

Similarly, the derivative with respect to $H_{kj}$ is:

{\small
\begin{equation*}
\frac{\partial L}{\partial H_{kj}} = \frac{\partial D_{KL}(V \parallel WH)}{\partial H_{kj}} + \mu_{k} \frac{\partial g_{2,k}(H)}{\partial H_{kj}} 
\end{equation*}
}

Inserting the derivatives from Eqs.~\eqref{eq:klH} and \eqref{eq:g2H}, we can write:

{\small
\begin{multline}
\frac{\partial L}{\partial H_{kj}} = \sum_{i'}\left(W_{i'k} - \frac{V_{i'j} W_{i'k}}{(WH)_{i'j}}\right) + \\ \mu_{k} \times
\begin{cases}
-\frac{\text{Den}_k - \text{Num}_k}{(\text{Den}_k)^2} & \text{if } k \in S_{MH}, j \in SI \\
\frac{\text{Num}_k}{(\text{Den}_k)^2} & \text{if } k \in S_{MH}, j \notin SI \\
0 & \text{otherwise}
\end{cases}
\end{multline}
}

\subsection{\texorpdfstring{Multiplicative Update Rule for $W_{ik}$}{Multiplicative Update Rule for W\_ik}}
\label{sec:multiplicative_update_for_W}

To optimize the objective under our constraints, we will update $H_{kj}$ and $W_{ik}$ according to the gradients we derived. In this section, we first derive the multiplicative update rule for $W_{ik}$.

We set the gradient to zero to satisfy the KKT condition:

{\small
\begin{multline*}
\frac{\partial L}{\partial W_{ik}} = 0 \Rightarrow \sum_{j'} H_{kj'} \left( 1 - \frac{V_{ij'}}{(WH)_{ij'}} \right) + \\ \lambda_{ik} \frac{\partial g_{1,ik}(W)}{\partial W_{ik}} = 0 \;.
\end{multline*}
}

Substituting the derivative of \( g_{1,ik}(W) \) we get:

{\small
\begin{multline*}
\sum_{j'} H_{kj'} \left( 1 - \frac{V_{ij'}}{(WH)_{ij'}} \right) 
= \\ -\lambda_{ik} \cdot
\begin{cases} 
1 & \text{if } i \in I_0,\;k \in S_{MH} \\
0 & \text{otherwise}
\end{cases}
\end{multline*}
}

We now rearrange the terms:

{\small
\begin{multline*}
\sum_{j'} H_{kj'} - \sum_{j'} H_{kj'} \frac{V_{ij'}}{(WH)_{ij'}} = 
\\ -\lambda_{ik} \cdot
\begin{cases} 
1 & \text{if } i \in I_0,\;k \in S_{MH}\\
0 & \text{otherwise}
\end{cases}
\end{multline*}
}

This can be further rearranged as

{\small
\begin{equation*}
\sum_{j'} H_{kj'} \frac{V_{ij'}}{(WH)_{ij'}} = \sum_{j'} H_{kj'} + \lambda_{ik}
\delta_{i \in I_0,\;k \in S_{MH}}
\end{equation*}
}

where $\delta_{i \in I_0,\;k \in S_{MH}}=1$ if $i \in I_0$ and $k \in S_{MH}$ and zero otherwise.
The above then yields

{\small
\begin{equation*}
\frac{\sum_{j'} H_{kj'} \frac{V_{ij'}}{(WH)_{ij'}}}{\sum_{j'} H_{kj'} + \lambda_{ik}
\delta_{i \in I_0,\;k \in S_{MH}}} = 1 \;.
\end{equation*}
}

The multiplicative update rule for \( W_{ik} \) can thus be written as:

{\small
\begin{equation*}
W_{ik} \leftarrow W_{ik}  \cdot \frac{\sum_{j'} H_{kj'} \frac{V_{ij'}}{(WH)_{ij'}}}{\sum_{j'} H_{kj'} +  \lambda_{ik} \delta_{i \in I_0,\;k \in S_{MH}}}
\end{equation*}
}

In the multiplier on the right-hand side, the numerator contains the terms that had a negative sign in the gradient of the Lagrangian, i.e., the terms that would have a positive sign when moving in the opposite direction of the gradient to minimize the Lagrangian. Similarly, the terms in the denominator are the ones that had a positive sign in the gradient, i.e., the terms that would have a negative sign when moving in the opposite direction of the gradient.

Therefore the numerator \( \sum_{j'} H_{kj'} \frac{V_{ij'}}{(WH)_{ij'}} \) corresponds to the positive part of the gradient during optimization, which pulls \( W_{ik} \) toward reducing the reconstruction error.
The denominator \( \sum_{j'} H_{kj'} + \lambda_{ik} \delta_{i \in I_0,\;k \in S_{MH}} \) includes the regularization term that controls the magnitude of \( W_{ik} \).

It is easy to see that the update rule maintains the nonnegativity of $W_{ik}$ as all terms that multiply $W_{ik}$ on the right-hand side of the rule are nonnegative.

%

\subsection{\texorpdfstring{Multiplicative Update Rule for $H_{kj}$}{Multiplicative Update Rule for H\_kj}}
\label{sec:multiplicative_update_for_H}

The update rule for $H_{kj}$ is derived from setting the gradient of the Lagrangian with respect to $H_{kj}$ to zero.
Given the expression

{\small
\begin{equation*}
\frac{\partial L}{\partial H_{kj}} = \sum_{i'} W_{i'k} \left( 1 - \frac{V_{i'j}}{(WH)_{i'j}} \right) + \mu_{k} \frac{\partial g_{2,k}(H)}{\partial H_{kj}} = 0
\end{equation*}
}

and inserting the gradient of the constraint, this becomes

{\small
\begin{multline*}
\frac{\partial L}{\partial H_{kj}} = \sum_{i'} W_{i'k} \left( 1 - \frac{V_{i'j}}{(WH)_{i'j}} \right) + \\ \left(\mu_{k}  
\cdot
\begin{cases}
-\frac{\text{Den}_k - \text{Num}_k}{(\text{Den}_k)^2} & \text{if } k \in S_{MH},\;j \in SI,  \\
\frac{\text{Num}_k}{(\text{Den}_k)^2} & \text{if } k \in S_{MH},\;j \notin SI\\
0 & \text{otherwise}
\end{cases}\right)
= 0
\end{multline*}
}

which can be simplified as

{\small
\begin{multline}
\frac{\partial L}{\partial H_{kj}} = \sum_{i'} W_{i'k} \left( 1 - \frac{V_{i'j}}{(WH)_{i'j}} \right) + \\ \mu_{k}  
\cdot
\delta_{k \in S_{MH}}\left(\frac{\text{Num}_k}{(\text{Den}_k)^2} -\delta_{j \in SI}\frac{1}{\text{Den}_k}\right)
= 0 \;.
\label{eq:lagrange_gradient_hkj_zero}
\end{multline}
}

We next derive two versions of the update rule.

\textbf{Version 1.} Starting from Eq.~\eqref{eq:lagrange_gradient_hkj_zero} we can rearrange the terms
as

{\small
\begin{multline*}
\sum_{i'} W_{i'k} \frac{V_{i'j}}{(WH)_{i'j}}
= \sum_{i'} W_{i'k} 
+ \\ \mu_{k}  
\cdot
\delta_{k \in S_{MH}}\left(\frac{\text{Num}_k}{(\text{Den}_k)^2} -\delta_{j \in SI}\frac{1}{\text{Den}_k}\right)
\end{multline*}
}

which then yields

{\small
\begin{equation*}
\frac{\sum_{i'} W_{i'k} \frac{V_{i'j}}{(WH)_{i'j}}}
{\sum_{i'} W_{i'k} 
+ \mu_{k}  
\cdot
\delta_{k \in S_{MH}}\left(\frac{\text{Num}_k}{(\text{Den}_k)^2} -\delta_{j \in SI}\frac{1}{\text{Den}_k}\right)} \;.
\end{equation*}
}

From this, we get the update rule for $H_{kj}$:


\begin{equation}
\scalebox{0.95}{$
H_{kj} \leftarrow H_{kj} \cdot
\frac{\sum_{i'} W_{i'k} \frac{V_{i'j}}{(WH)_{i'j}}}
{\sum_{i'} W_{i'k} 
+ \mu_{k}  
\cdot
\delta_{k \in S_{MH}}\left(\frac{\text{Num}_k}{(\text{Den}_k)^2} -\delta_{j \in SI}\frac{1}{\text{Den}_k}\right)} $}
\label{eq:update_rule_for_H_version1}
\end{equation}

The update rule in Eq.~\eqref{eq:update_rule_for_H_version1} above is appealing due to its symmetrical form to the update rule of $W_{ik}$. In the multiplier on the right-hand side the numerator is always nonnegative and the denominator is nonnegative if $\sum_{i'} W_{i'k}$ is larger than the term with $\mu_k$. In our experiments, the denominator has consistently remained nonnegative and hence the updates maintain nonnegativity of $H$. Thus we use this update rule due to its appealing symmetry. However, it is also possible to use an alternate update rule with guaranteed nonnegativity and we derive it below.

\textbf{Version 2.} Starting from Eq.~\eqref{eq:lagrange_gradient_hkj_zero} we can rearrange the 
terms as

{\small
\begin{multline*}
\sum_{i'} W_{i'k} \frac{V_{i'j}}{(WH)_{i'j}}
+ \mu_k \cdot \delta_{k \in S_{MH}} \delta_{j \in SI}\frac{1}{\text{Den}_k}
= \\ \sum_{i'} W_{i'k} 
+ \mu_k \cdot \delta_{k \in S_{MH}} \frac{\text{Num}_k}{(\text{Den}_k)^2}
\end{multline*}
}

This then yields

{\small
\begin{equation*}
\frac{\sum_{i'} W_{i'k} \frac{V_{i'j}}{(WH)_{i'j}}
+ \mu_k \cdot \delta_{k \in S_{MH}} \delta_{j \in SI}\frac{1}{\text{Den}_k}
}
{\sum_{i'} W_{i'k} 
+ \mu_k \cdot \delta_{k \in S_{MH}} \frac{\text{Num}_k}{(\text{Den}_k)^2}}
= 1 \;.
\end{equation*}
}

The multiplicative update for $H_{kj}$ then becomes

{\small
\begin{equation}
H_{kj}
\leftarrow
H_{kj}
\cdot
\frac{\sum_{i'} W_{i'k} \frac{V_{i'j}}{(WH)_{i'j}}
+ \mu_k \cdot \delta_{k \in S_{MH}} \delta_{j \in SI}\frac{1}{\text{Den}_k}
}
{\sum_{i'} W_{i'k} 
+ \mu_k \cdot \delta_{k \in S_{MH}} \frac{\text{Num}_k}{(\text{Den}_k)^2}}
\label{eq:update_rule_for_H_version2}
\end{equation}
}

The update rule in Eq.~\eqref{eq:update_rule_for_H_version2}
always maintains the nonnegativity of $H_{kj}$ as all terms in the multiplier are nonnegative. Therefore this rule can be used as an alternative to the first update rule (Eq.~\eqref{eq:update_rule_for_H_version1}) if any situation arises where Eq.~\eqref{eq:update_rule_for_H_version1} would not maintain nonnegativity, e.g. if the multiplier $\mu_k$ were set to a very large value. In practice, we have found Eq.~\eqref{eq:update_rule_for_H_version1} to work well across all our experiments but we provide this alternate update rule for completeness.

\subsection{Details of the Datasets}
\label{sec:details_of_the_datasets}

We discuss the formation of our case study data set and synthetic data set below.

\textbf{Real data set.}
Our real data, discussed in Section~\ref{sec:dataset} of the main paper, concerns YouTube discussion of viewers for vlogs of Finnish YouTubers and relates to an ongoing study of peer mental health support. YouTube is a massive video sharing platform where many YouTubers have risen to prominence as influencers. The focus of interest in this data domain is the viewers' discussion of mental health aspects, rather than all discussion, hence the domain is a suitable case study for topic modeling of minority topics. To form the data set, we began by selecting 20 pre-identified Finnish YouTubers from our public health experts, who focus primarily on vlogs related to topics such as mental health issues, targeting a younger audience. 
The topics found by our proposed method from this real data are discussed later in this Appendix in Section \ref{sec:discovered_topics}.

\textbf{Synthetic data set.} We also formed a synthetic data set where we started from the real data set and injected known mental health content as ground truth topical contents, as described below. This data set was used to compare the performance of several methods in modeling the ground truth topical content in Sections \ref{sec:evaluation} and \ref{sec:findings} of the main paper.

To build our synthetic dataset, we injected topic-related words into randomly selected sentences to simulate mental health discussions. We began by defining a synthetic ground truth as 18 mental health topics to be injected, each with a collection of related Finnish words. This synthetic ground truth was designed to be realistic for the mental health case.

To generate the synthetic data we first chose 500 documents (comment sentences under YouTube vlogs) from our dataset at random. Each document then had a probability of 10 percent to be injected with additional words relating to mental health. When choosing a document, we selected one of the synthetic mental health topics randomly and inserted four of its words in random spots into the document. As a result, some documents have been injected with a relevant mental-health topic and the label of the injected topic is recorded for each document. Note that mental health content was injected into a minority of documents and forms a minority of words in those documents. The majority of the 500 sentences were left unmodified and were given a ground-truth label ``-1" indicating they do not contain injected mental health topics. This method allowed us to create a realistic synthetic dataset by combining regular documents with topic-specific terms.

Since we know the ground-truth injected mental health topic (if any) for each of the 500 documents, we can use them for performance evaluation. We use clustering quality measures for this purpose: given a topic modeling result by one of the compared methods, we assign each document to its estimated majority topic as a cluster label. The estimated cluster labels are then evaluated against the ground truth labels described above, by the well-known clustering quality measures normalized mutual information (NMI) and by the Purity score.

Note that for the Purity score in Section \ref{eq:purity_score} (which estimates what proportion of all documents have the same ground-truth label as the majority label of their assigned cluster), since we are only interested in mental health related documents we pick the majority class of each cluster from its mental health injected documents, and ignore any clusters having no mental health injected documents. This rewards clusterings where non-mental health documents are placed into separate clusters from mental health injected ones.

\subsection{Details of the hyper-parameter tuning, computing systems, and time complexity}
\label{sec:hyperparameter_tuning_details}

For all the compared SOTA baselines, we sought the same number of topics and ensured fair comparison with equal hyperparameter configurations. For NMF, we used default settings: \(\alpha_W = 0.0\), \(\alpha_H = \text{'same'}\), and \(l1\_ratio = 0.0\). We tested both solvers and observed no substantial differences in performance. For LDA, we used the default priors: \texttt{doc\_topic\_prior} \(= 1 / \texttt{n\_components}\) and \texttt{topic\_word\_prior} \(= 1 / \texttt{n\_components}\). We found that varying these priors did not significantly affect model performance. For Top2Vec, we followed the authors' recommended setting, as these were optimized for best performance. SeededLDA was evaluated with three parameter configurations: the default settings suggested by its authors and two additional random samples around the defaults. For the Anchored CoreX method, the key hyperparameter was anchor strength, which we tested across three reasonable values, selecting the best-performing value \(4\) for comparison. Our method similarly used three configurations, ensuring no method had an unfair advantage. We used publicly available baseline codes; including NMF \footnote{\url{https://scikit-learn.org/dev/modules/generated/sklearn.decomposition.NMF.html}}, LDA \footnote{\url{https://scikit-learn.org/1.5/modules/generated/sklearn.decomposition.LatentDirichletAllocation.html}}, Top2Vec \footnote{\url{https://github.com/ddangelov/Top2Vec}}, KeyATM \footnote{\url{https://keyatm.github.io/keyATM/}}, GuidedNMF \footnote{\url{https://github.com/jvendrow/GuidedNMF}}, SeededLDA \footnote{\url{https://github.com/bsou/cl2_project/tree/master/SeededLDA}}, Corex \footnote{\url{https://github.com/gregversteeg/corex_topic}}, GuidedLDA \footnote{\url{https://github.com/vi3k6i5/GuidedLDA}}, ProdLDA \footnote{\url{https://pyro.ai/examples/prodlda.html}}, BERTopic \footnote{\url{https://maartengr.github.io/BERTopic/index.html}}, and FASTopic \footnote{\url{https://github.com/BobXWu/FASTopic}}.

For our experiments, we used two alternative computing systems. We processed 150k data points using a local workstation with 32GB of RAM and an NVIDIA RTX 2000 Ada GPU. We used a High-Performance Computing (HPC) on the web interface to analyze the entire dataset, which has 4.5 million comments. 

Regarding time complexity, since the core of our constrained NMF technique is matrix operations, time complexity increases linearly with the number of features (rows of matrix \(W\)) and data points (columns of matrix \(V\)). In particular, \(O(T\cdot M N K)\) is the time complexity per iteration, where \(M\) is the number of rows in \(V\), \(K\) is the number of latent features, and \(N\) is the number of data points. The computational cost grows with dataset size due to the matrix multiplications needed, especially when \(N\) increases. Furthermore, the total computation time may be affected by the growth of \(T\), the number of iterations needed for convergence. 

\subsection{Training Details}
\label{sec:training_details}
Pre-processing the data involved eliminating symbols and stopwords from the NLTK\footnote{\url{https://www.nltk.org/}} stopwords list and a custom Finnish stopwords list\footnote{\url{https://github.com/stopwords-iso/stopwords-fi}}. The Voikko library\footnote{\url{https://github.com/voikko/corevoikko}} for spellchecking and stemming was then used to reduce inflected words to their simplest versions which is required particularly for Finnish language. We used Term Frequency-Inverse Document Frequency (TF-IDF) \cite{SALTON1988513} transformation with \texttt{max\_df= 0.9} and \texttt{min\_df= 0.2} for the real dataset, but no \texttt{max\_df} or \texttt{min\_df} values were used for the synthetic dataset. In our model, the parameter \texorpdfstring{$\theta_{\text{min}}$}{theta\_min}= 0.4, and one-third of the total topics were determined as mental health-related. The model demonstrated rapid improvement, dropping significantly in the first 20 iterations, followed by minor changes.  The learning rate $\eta = 0.001$ was used to update the Lagrange multipliers $\lambda$ and $\mu$. We found $W_{\text{max}} = 1 \times 10^{-9}$ to be optimal across trials. Our code is available\footnote{\url{https://github.com/anonymous-researcher-573/Contrained_non-negative-matrix-factorization}}.

\subsection{Topics Discovered by the Proposed Method}
\label{sec:discovered_topics}
We display the themes found by our Constrained NMF model applied to both synthetic and real-world datasets respectively in Tables~\ref{tab:tableA6} and Tables~\ref{tab:tableA2}--\ref{tab:tableA5}. These tables demonstrate how well the methodology can yield insightful and superior mental health-related conversation topics from adolescent peer support discussions. Tables ~\ref{tab:tableA2}--\ref{tab:tableA5} in 4 parts, demonstrates the topics discovered in the real dataset, while Table~\ref{tab:tableA6} displays those discovered in our synthetic dataset. Each table contains the top ten terms for each topic in both Finnish and English translation. 

\textbf{Topics Discovered in the Real Dataset.}
Table~\ref{tab:tableA2} lists first 15 mental health-related topics, with the rest in Tables~\ref{tab:tableA3}--\ref{tab:tableA5} being non-mental health topics. For each topic, the words are listed twice: first as the original Finnish words and then as English translations.
The topics are interpretable, when analyzed together with their top documents, as is common in topic modeling.

Topic 1 features odotus (expectation), joulu (Christmas) and mielenterveysongelma (mental health problem) as top words, this is because its top documents feature discussion of expectation, such as expectation of Christmas, but also comments such as “expectation versus reality”, as well as low assessments of the future such as seeing mental health problems, intoxicants, and grudges being issues.
Similarly, Topic 5 is on relationships in families, and words like perhe (family), isä (father), äiti (mother), and lapsi (child) appear prominently. Topic 6 focuses on mental health treatment and support, including terminology such as hoito (therapy), tuki (support), and ahdistus (anxiety). Topic 4 focuses on trauma and psychosis, using terms such as trauma (trauma) and psyko (psycho).
Topic 7 delves into psychosis, trust, and emotional issues which reflects trust issues in relationships and the impact of psychosis on mental well-being:
the topic features top documents having discussion of trust and critical attitude towards various sources of information (internet, celebrities, research, politicians, cults), and betrayal of trust by frauds, and even trusting computer security; on the other hand there was also a comment suggesting one would rather bear pain at home rather than going to a hospital, connecting pain to issues of trust.
The remaining subjects cover a variety of mental health issues, including self-harm (Topic 0), schizophrenia (Topic 9), and interactions with healthcare providers (Topic 14), demonstrating the model's ability to represent varied topics related to mental health.

\textbf{Topics Discovered in the Synthetic Dataset.}
Table~\ref{tab:tableA6} highlights the topics observed in the synthetic dataset, which help to validate the efficiency of our model. Again, for each topic, the words are listed twice: first as the original Finnish words and then as English translations. We specified the total number of topics as well as the mental health issues. Table~\ref{tab:tableA6} has 7 mental health issues (1-7), and the remaining topics  from 8 to 20 are non-mental health. For example, Topic 1 includes words like väkivalta (violence), psykoosi (psychosis), and skitsofrenia (schizophrenia), whereas Topic 2 covers topics about masennus (depression), hoito (treatment), and vertaistuki (peer support). Topic 4 addresses broader mental health issues, including persoonallisuushäiriö (personality disorder) and emotional struggles.
Our model's consistent understanding of family, trauma, treatment, and support tendencies across both datasets highlights its effectiveness in discovering high-quality mental health topics. 
Note that in this synthetic data the mental health content was injected from synthetic original ground truth mental health topics. The topics found by the model correspond well to the original ground truth topics. Note that the quality of the model was also quantitatively shown by its best clustering performance out of all compared models in Section \ref{sec:findings} of the main paper.

\begin{table*}[t]
    \centering
    \begin{small}
    \begin{tabular}{lp{0.8\linewidth}}
    \hline
    \textbf{Topic} & \textbf{Top 10 Topic Words} \\
    \hline
    Topic 0 & hullu, ulkonäkö, lääke, lehti, ihminen, yrittää, itsemurha, mieli, keskustelu, mies \\
         & crazy, appearance, medicine, magazine, human, try, suicide, mind, conversation, man \\
    \hline
    Topic 1 & odotus, joulu, mielenterveysongelma, ihminen, pettymys, tunnistaa, pelko, autismi, yksinäisyys \\
    & expectation, christmas, mental health problem, human, disappointment, recognize, fear, autism, loneliness \\
    \hline
    Topic 2 & surullinen, itsemurha, ihminen, kärsimys, kuolla, metsä, jutella, johtaa, saada, keskustelu \\
     & sad, suicide, human, suffering, die, forest, chat, lead, get to, conversation \\
    \hline
    Topic 3 & tukea, adhd, kriisi, onnellisuus, itsemurha, suomi, julkisuus, tieto, addiktio, anna \\
     & to support, adhd, crisis, happiness, suicide, finland, publicity, knowledge, addiction, to give \\
    \hline
    Topic 4 & trauma, psyko, ihminen, empatia, keskustelu, väkivalta, pitää, saada, yhteiskunta, tsemppiä \\
     & trauma, psycho, human, empathy, conversation, violence, have to, get to, society, rooting for you \\
    \hline
    Topic 5 & perhe, isä, äiti, lapsi, vanha, koti, yhteisö, vuotiaana, lestadiolainen, tappaa \\
     & family, father, mother, child, old, home, community, years old, laestadian, to kill \\
    \hline
    Topic 6 & hoito, tuki, ahdistus, vihapuhe, mieli, henkilö, kipu, jakaa, saada, oire \\
    & treatment, support, anxiety, hate speech, mind, person, pain, share, get to, symptom \\
    \hline
    Topic 7 & luottaa, kipu, psykoosi, hallusinaatio, ongelma, mielenterveys, paha, tuntea, tosi, keskustelu \\
     & trust, pain, psychosis, hallucination, problem, mental health, bad, feel, real, conversation \\
    \hline
    Topic 8 & itku, kannabis, potilas, saada, arvo, tukea, ilo, mieli, tasa, israel \\
     & crying, cannabis, patient, get to, value, to support, joy, mind, even, israel \\
    \hline
    Topic 9 & pelko, neuvo, jutella, masennus, skitsofrenia, lapsi, ihminen, keskustelu, tunne, saada \\
    & fear, advice, to chat, depression, schizophrenia, child, human, conversation, feeling, get to \\
    \hline
    Topic 10 & huume, poliisi, suomi, jengi, väkivalta, rikollinen, käyttäjä, saada, itsetunto, nykyään \\
     & drug, police, finland, gang, violence, criminal, user, get to, self confidence, nowadays \\
    \hline
    Topic 11 & viha, keskustelu, mielenterveys, motivaatio, tukea, itsemurha, väkivalta, ihminen, aihe, pelko \\
    & hate, conversation, mental health, motivation, to support, suicide, violence, human, topic, fear \\
    \hline
    Topic 12 & tunne, ilo, huume, raamattu, rakkaus, ihminen, jumala, stressi, kestää, rauha \\
     & feeling, joy, drug, bible, love, human, god, stress, bear it, peace \\
    \hline
    Topic 13 & terapia, keskustelu, vakava, ihminen, tuomita, väite, tuntea, pelastua, puhua, paine\\
    & therapy, conversation, serious, human, sentence, claim, feel, be saved, speak, pressure \\
    \hline
    Topic 14 & lääkäri, ärsyttävä, masennus, paniikki, häiriö, rutto, syy, jaksaa, autismi, saada \\
     & doctor, annoying, depression, panic, disorder, plague, reason, bear to, autism, get to \\
    \hline
    Topic 15 & pelottava, tehä, mulle, vois, vittu, jeesus, vesi, israel, uskaltaa, uida \\
     & scary, to do, to me, one could, fuck, jesus, water, israel, dare, to swim \\
    \hline
    Topic 16 & tietää, ottaa, lukea, kysymys, odottaa, tosta, vastata, totuus, pelottaa, mr \\
    & know, take, read, question, wait, from there, answer, truth, be scared, mr \\
    \hline
    Topic 17 & mun, kaveri, vaa, lähteä, pelata, mieli, seuraavaks, selittää, poi, paa \\
    & my, friend, just, leave, play, mind, next, explain, away, to put \\
    \hline
    Topic 18 & vaarallinen, mäkipelto, kuulla, määrä, homma, nuori, meinata, minecraft, sota, tsemppiä \\
    & dangerous, youtuber's name, hear, amount, matter, young, to mean, minecraft, war, rooting for you \\
    \hline
    Topic 19 & kiva, venäjä, sä, salaliittoteoria, metsä, mainita, sota, mainos, jakso, sua \\
    & nice, russia, you, conspiracy theory, forest, mention, war, ad, episode, of you \\
    \hline
    Topic 20 & kohta, kommentti, katto, hieno, maa, tosi, like, paljo, olo, planeetta \\
    & soon, comment, ceiling, nice, earth, really, near, much, feeling, planet 
    \end{tabular}
    \end{small}
\caption{Topics discovered by our model in the real dataset, part 1, topics 0-20.}
\label{tab:tableA2}
\end{table*}

\begin{table*}[t]
\centering
\begin{small}
\begin{tabular}{lp{0.8\textwidth}}
\hline
\textbf{Topic} & \textbf{Top Topic Words} \\
\hline
Topic 21 & mahtava, asu, seuraava, kaupunki, tubettaja, ehdoton, elokuva, supo, helsinki, mielenkiintonen, pitänyt, huippu, vaatia, polttaa \\
& great, outfit, next, city, youtuber, absolute, movie, security and intelligence service, helsinki, interesting, should have, top, demand, burn \\
\hline
Topic 22 & kanava, nähdä, tykätä, kiinnostava, alkaa, alku, selvä, raha, toivottava, kuullu, löysä, tilaus, artolauri \\
& channel, see, like, interesting, begin, beginning, clear, money, hopeful, heard, loose, order, artolauri (personname)\\
\hline
Topic 23 & tarina, historia, sana, taitaa, löytyä, laki, tärkeä, käynyt, loistava, asua, legenda, netti, tapahtunut, pitää, rakentaa, tarkoitus, sattua, mieli \\
& story, history, word, be likely, find, law, imporant, happened, brilliant, live, legend, internet, happened, must, build, purpose, hurt, mind \\
\hline
Topic 24 & saada, kuu, tossa, huomata, ostaa, lähde, jatko, seurata, liikkua, kuula, selkeä, oikeus, tapahtuma, pyytää, peru, selvitä \\
& get, month, there, notice, buy, source, continuation, follow, move, ball, clear, justice, event, request, cancel, survive \\
\hline
Topic 25 & katsoa, kiinnostaa, elämä, vähä, jaksaa, pari, äiti, uus, supervoima, ulos, suku, elää, sopimus, vapaa \\
& see, interest, life, little, to bear, couple, mother, new, superpower, out, family, live, contract, free \\
\hline
Topic 26 & fakta, onni, ihmeellinen, outo, kissa, kuulostaa, koira, väärä, rakastaa, ase, musiikki, joulu, mite, asukas, onnettomuus, syntyä \\
& fact, happiness, wonderful, strange, cat, sound like, dog, wrong, love, weapon, music, christmas, how, tenant, accident, be born \\
\hline
Topic 27 & osata, miettiä, kuolema, ruotsi, sitte, naapuri, mysteeri, siisti, kylmä, hahmo, sanottu, kestää, luokka, laulu \\
& know how to, think, death, sweden, then, neighbor, mystery, clean, cold, character, said, to bear, class, song \\
\hline
Topic 28 & aihe, mielenkiintoinen, tehty, petteri, kans, lopettaa, väärin, tehny, mieli, mikkonen, halloween, katsonut, tuleva, mahtava, saanu \\
& topic, interesting, done, peter, also, stop, wrong, done, mind, mikkonen (surname), halloween, watched, future, great, gotten \\
\hline
Topic 29 & ihminen, top, lapsi, eläin, ymmärtää, jäädä, vaikuttaa, tyhmä, syödä, tappaa, luonto, koe, käyttö, tuntea, lukenut, tarvita \\
& human, top, child, animal, understand, stay, influence, stupid, eat, kill, nature, experiment, use, feel, read, need \\
\hline
Topic 30 & työ, koulu, oikeesti, henkilö, musta, vika, jenna, käsi, talo, vitsi, tieten, numero, kauhu, milkyllänaa, kallio, puhe, haluinen \\
& work, school, really, person, black, flaw, jenna, hand, house, joke, of course, number, horror, milkyllänaa, rock, speech, wanting \\
\hline
    \end{tabular}
    \end{small}
\caption{Topics discovered by our model in the real dataset, part 2, topics 21-30.}
\label{tab:tableA3}
\end{table*}

\begin{table*}[t]
\centering
\begin{small}
\begin{tabular}{lp{0.8\textwidth}}
\hline
\textbf{Topic} & \textbf{Top Topic Words} \\
\hline
Topic 31 & maailma, suomi, peli, tieto, löytää, rikas, alue, gta, keksi, kuoltu, ykkönen, poistaa, korona, rakennus, auttaa, ikä, rokote, synty \\
& world, finland, game, knowledge, find, rich, area, gta, cookie, died, number one, remove, corona, building, help, age, vaccine, birth \\
\hline
Topic 32 & muistaa, paikka, ajatella, haluta, pystyä, käärme, jättää, onks, päättää, nähdä, todistaa, valo, kuunnella, raamattu, ero, mahdoton \\
& remember, place, think, want, be able to, snake, leave, is it, decide, see, prove, light, listen, bible, difference, impossible \\
\hline
Topic 33 & paha, mielenkiintoinen, vastaus, avaruus, katsoma, eurooppa, suositella, paikko, maa, samanlainen, vanha, suomi, vankila, välttää, etsiä, sydän, todiste \\
& evil, interesting, answer, space, looking, europe, recommend, replacement, country, similar, old, finland, prison, avoid, seek, heart, proof \\
\hline
Topic 34 & tuoda, vanha, yö, jännä, silmä, kuuluu, uni, nään, isä, herätä, normaali, tie, ihmetellä, aamu, kommentoida, veli \\
& bring, old, night, exciting, eye, belong, dream, i see, father, wake, normal, road, wonder, morning, comment, brother \\
\hline
Topic 35 & juttu, presidentti, iso, veikata, suomi, japani, setti, sopia, turvallinen, mäki, isojalka, miljardi, murha \\
& story, president, big, wager, finland, japan, set, fit, safe, hill, bigfoot, billion, murder \\
\hline
Topic 36 & ääni, nauraa, nähny, mieli, puuttua, tapaus, ihmeellinen, lahko, kuollut, mahdollinen, tulo, kiinni, nähä, rikollinen, alue, sotilas, esine \\
& sound, laugh, seen, mind, lack, event, wondrous, cult, dead, possible, product, caught, see, criminal, area, soldier, object \\
\hline
Topic 37 & nimi, kyl, loppu, mielenkiintoisa, pitkä, katto, näkyä, tuttu, poika, loppua, tutkia, matka, nähnyt, leffa, huone, lemppari \\
& name, yes, end, interesting, long, roof, see, familiar, son, end, investigate, trip, seen, movie, room, favorite \\
\hline
Topic 38 & luulla, laittaa, sisältö, tullu, kattonu, päin, upea, pistää, valtio, dinosaurus, teko, raja, asiallinen, tylsä \\
& think, to put, contents, has come, seen, toward, gorgeous, to stick, government, dinosaur, action, limit, reasonable, boring \\
\hline
Topic 39 & uskoa, jumala, puhua, liittyä, ikinä, ihminen, uskonto, syy, paska, tulevaisuus, kirjoittaa, kieli, millon, ihmiskunta, tuhota, kanta, kokea, raamattu \\
& believe, god, speak, join, ever, human, religion, reason, shit, future, write, language, when, humanity, destroy, stance, experience, bible \\
\hline
Topic 40 & jengi, pää, viikko, iso, voima, unohtaa, virhe, onnee, sais, jalka, porukka, sahein, fiksu, kuulemma, tienny, keksintö, päässyt, hinta \\
& gang, head, week, big, power, forget, mistake, congratulations, could get, foot, crowd, neat, smart, as i heard, known, invention, gotten, price \\
\hline
    \end{tabular}
    \end{small}
\caption{Topics discovered by our model in the real dataset, part 3, topics 31-40.}
\label{tab:tableA4}
\end{table*}

\begin{table*}[t]
\centering
\begin{small}
\begin{tabular}{lp{0.8\textwidth}}
\hline
\textbf{Topic} & \textbf{Top Topic Words} \\
\hline
Topic 41 & suomalainen, mies, kertoa, kova, nainen, hauska, kirja, teoria, kuultu, helvetti, tapa, tarkka, kappale, mieli - liittyvä, linna, pitää, finntop \\
& finnish, man, tell, hard, woman, funny, book, theory, heard, hell, manner, precise, piece, mind, related, castle, keep, finntop \\
\hline
Topic 42 & kattoo, kerto, kuolla, salainen, itsenäisyyspäivä, käyny, saatana, onneks, ohjelma, rauha, toimia, puoli, kiehtova, vaimo, saada, jätkä \\
& see, tell, die, secret, independence day, visited, satan, luckily, program, peace, act, half, fascinatinng, wife, get, dude \\
\hline
Topic 43 & pakko, oppia, ai, elää, aivo, toivoa, ajatus, turha, herra, muuttua, myöhä, itä, kauhea, ansainnut, nostaa, laskea, kasvi \\
& need, learn, oh, live, brain, hope, thought, futile, sir, change, late, east, horrible, deserved, raise, lower, plant \\
\hline
Topic 44 & idea, luoja, ongelma, luu, poliisi, pallo, amerikka, värinen, salaliitto, laita, huomio, yllättää, tekemä \\
& idea, creator, problem, bone, police, ball, america, colored, conspiracy, put, attention, surprise, made by \\
\hline
Topic 45 & käyttää, laiva, riittää, uutinen, arvo, lisätä, mielenkiintosia, laulaa, ruoka, viisas, vaunu, upota, pohja \\
& use, ship, suffice, news, value, add, interesting, sing, food, wise, wagon, sink, bottom \\
\hline
Topic 46 & mainittu, sarja, parka, arto, tavata, lauri, tori, käsitellä, ajaa, armeija, anna, oikeen, ihme, alotukset, aiheinen, positiivinen, merkitys \\
& mentioned, series, poor, arthur, meet, larry, market, deal with, drive, army, give, real, miracle, beginnings, topical, positive, meaning \\
\hline
Topic 47 & varma, kuva, näyttää, kokemus, mieli, jakaa, väittää, ilta, vidi, katsella, äijä, uskomaton, haamu, räjäyttää - kone \\
& sure, picture, show, experiance, mind, share, claim, evening, vid, watch, dude, incredible, ghost, explore, machine \\
\hline
Topic 48 & mennä, jatkaa, tilata, auto, malli, väli, kallis, ihme, super, meri, maksa, väri, salaisuus, niinkuin \\
& go, continue, order, car, model, distance, expensive, miracle, super, ocean, pay, color, secret, as in \\
\hline
Topic 49 & biisi, tilaaja, päästä, tehnyt, käydä, helppo, tiennyt, sairas, pelkkä, pitäs, vahva, tutkimus, paikata, ansaita, tajuta, kunta \\
& song, customer, make it, done, visit, easy, known, sick, mere, should, strong, research, compensate, earn, realize, municipality \\
\hline
\end{tabular}
\end{small}
\caption{Topics discovered by our model in the real dataset, part 4, topics 41-49.}
\label{tab:tableA5}
\end{table*}

\begin{table*}[t]
\centering
\begin{small} 
\begin{tabular}{lp{0.8\linewidth}}
\hline
\textbf{Topic} & \textbf{Top 10 Topic Words} \\
\hline
Topic 1 & väkivalta, kokemus, psykoosi, itsemurha, oire, syy, ihminen, suru, pöllö, skitsofrenia \\
       & violence, experience, psychosis, suicide, symptom, cause, human, sorrow, cuckoo, schizophrenia \\
\hline    
Topic 2 & kannabis, masennus, rentoutuminen, jumala, tuomita, hoito, hoitaja, kuunnella, jeesus, vertaistuki \\
       & cannabis, depression, relaxation, god, sentence, treatment, nurse, listen, jesus, peer support \\
\hline
Topic 3 & keskustelu, viha, tukea, ongelma, tunne, lapsi, ihminen, mulle, skitsofrenia, adhd \\
       & conversation, hate, to support, problem, feeling, child, human, to me, schizophrenia, adhd \\
\hline 
Topic 4 & mielenterveys, sairaala, sota, ihminen, tukea, hoito, tunne, kokea, persoonallisuushäiriö, uskoa \\
       & mental health, hospital, war, human, to support, treatment, feeling, to experience, personality disorder, to believe \\
\hline       
Topic 5 & uskoa, kokea, perhe, erottaa, sun, parisuhde, huolestua, alotuksista, pitäiskö, tukea \\
       & to believe, to experience, family, to separate, your, relationship, to get worried, of beginnings, should one, to support \\
\hline 
Topic 6 & neuvonta, perhe, kuunnella, psykoterapia, kannabis, hoito, skitsofreenikko, lauri, arto, hitto \\
       & counseling, family, listen, psychotherapy, cannabis, treatment, schizophrenic, larry, artie, damn \\
\hline 
Topic 7 & vihapuhe, ihminen, tuki, sairaala, päihde, kannabis, vertaistuki, totuus, suhun, voldemortilta \\
       & hate speech, human, support, hospital, intoxicant, cannabis, peer support, truth, to you, from voldemort \\
\hline 
Topic 8 & mielenkiintoinen, katsottava, oikeesti, aihe, vitsi, osata, pitäis, päästä, sul, toiminta \\
       & interesting, watchable, for real, topic, joke, be able, should, get to, with you, action \\
\hline 
Topic 9 & käydä, ny, hienostaa, venäjä, koulu, lauantai, selittää, ratkaisu, kpl, ku \\
       & visit, now, make fancy, russia, school, saturday, explain, solution, amount, when \\
\hline 
Topic 10 & kirja, paa, kiva, suomi, sääli, sisältö, seura, nii, puolustusvoima, eläin \\
        & book, put, nice, finland, pity, content, company, yeah, defence forces, animal \\
\hline 
Topic 11 & hieno, para, oot, tunnettu, maailma, onni, sit, kanava, suomi, maa \\
& great, poor, you are, known, world, happiness, then, channel, finland, country \\
\hline 
Topic 12 & malli, jatkaa, kuuluu, kieli, ranska, raha, nato, pitää, pelkkä, ruth \\
& model, continue, belong, language, france, money, nato, should, mere, ruth \\
\hline 
Topic 13 & mun, aihe, kova, tuoda, fakta, ois, mulla, kanava, muistaa, tilata \\
& mine, topic, hard, bring, fact, would be, i have, channel, remember, order \\
\hline 
Topic 14 & jes, homma, mun, pelottava, seuraava, sarja, fiilis, tieto, kuu, kanada \\
& yes, matter, mine, scary, next, series, feeling, knowledge, month, canada \\
\hline 
Topic 15 & sun, suomi, ruumis, mukava, chicago, oho, mies, kiehtovii, uudestaan, ladata \\
& your, finland, body, nice, chicago, wow, man, fascinating, again, load \\
\hline 
Topic 16 & tarina, lista, star, pyrkiä, poi, tieto, suosikki, ruusu, pystyä, suomi \\
& story, list, star, attempt, poi, knowledge, favorite, rose, be able to, finland \\
\hline 
Topic 17 & nato, uskoa, aloitus, viikinki, helvetti, tosi, suomi, liittymä, mua, nokia \\
& nato, believe, beginning, viking, hell, real, finland, cell phone plan, me, nokia \\
\hline 
Topic 18 & oot, sun, jee, mahtava, seuraavaks, lempi, videoo, tykätä, mona, katto \\
& you're, your, yeah, great, next, favorite, video, like, mona, ceiling \\
\hline 
Topic 19 & tapahtua, mieli, kiinnostava, selvä, illuusio, ihminen, päästä, mi, area, seleeni \\
& happen, mind, interesting, clear, illusion, human, to get to, mi, area, selenium \\
\hline 
Topic 20 & ihminen, epäillä, tarvita, oo, ukraina, elää, nähny, hari, kalifornia, kandassa \\
& human, doubt, need, oo, ukraine, live, seen, hari, california, in canada \\
\end{tabular}
\end{small}
\caption{Topics discovered by our model in the synthetic dataset.}
\label{tab:tableA6}
\end{table*}

\subsection{Error Bars of KL Divergence from Multiple Model Runs}
\label{sec:error_bars}
\begin{figure}[t]
\centering
\includegraphics[width=\linewidth]{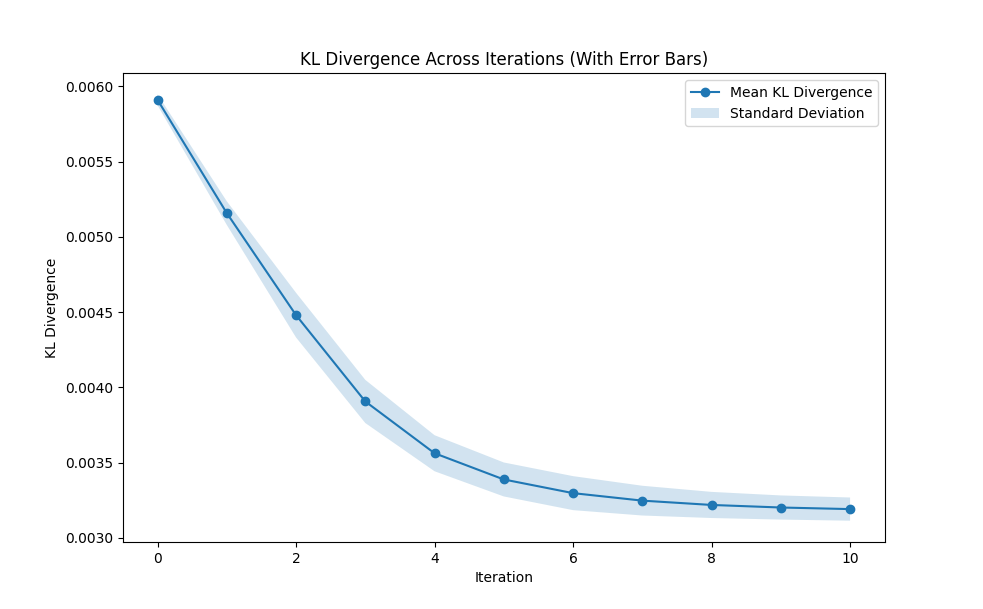}
\caption{KL Divergence Across Iterations with Error Bars.}
\label{fig:error_bar}
\vskip -0.1in
\end{figure}

In our research, 
the model was run using 10 different random seeds to ensure robustness and reduce biases in performance evaluation.
In each iteration, the elements of the matrices $V$ and $H$ were initialized as absolute values of normally-distributed zero-mean random numbers with standard deviation $0.01$.
For each iteration, we determined the mean KL divergence over initializations, which reflects the average divergence between the learned and target distributions and therefore serves as an important performance parameter. To measure the variability of KL divergence data, we calculated the standard deviation, which represents the degree of dispersion between values obtained from different trials. 
Figure \ref{fig:error_bar} shows the mean KL divergence (curve) and 10 times the standard deviation (error bars). It shows that the objective function stabilizes at a stationary point after decreasing monotonically, as shown by the mean curve. In line with the theoretical constraints of multiplicative update rules in non-convex optimization, we point out that this does not always imply convergence to a local minimum.

\begin{figure*}[!t]
\centering
\includegraphics[width=\linewidth]{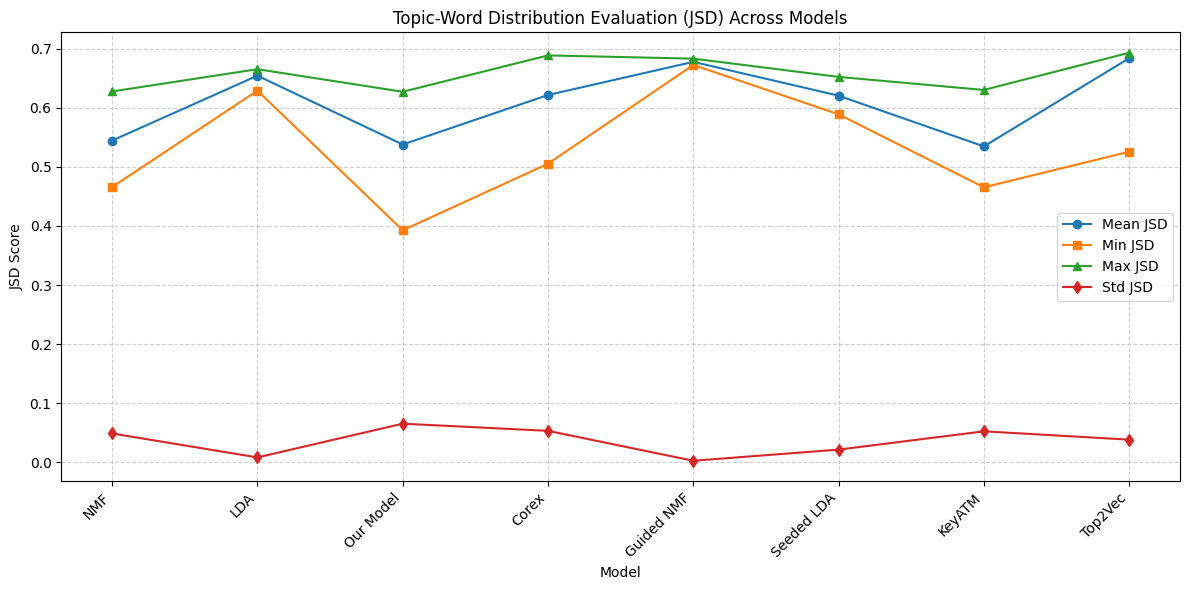}
\caption{Topic Quality using JSD Score}
\label{fig:JSD.png}
\end{figure*}

\section{Ablation Studies}
\label{appendix:sensitivity}
\begin{figure*}[!t]
\centering
\includegraphics[width=\linewidth]{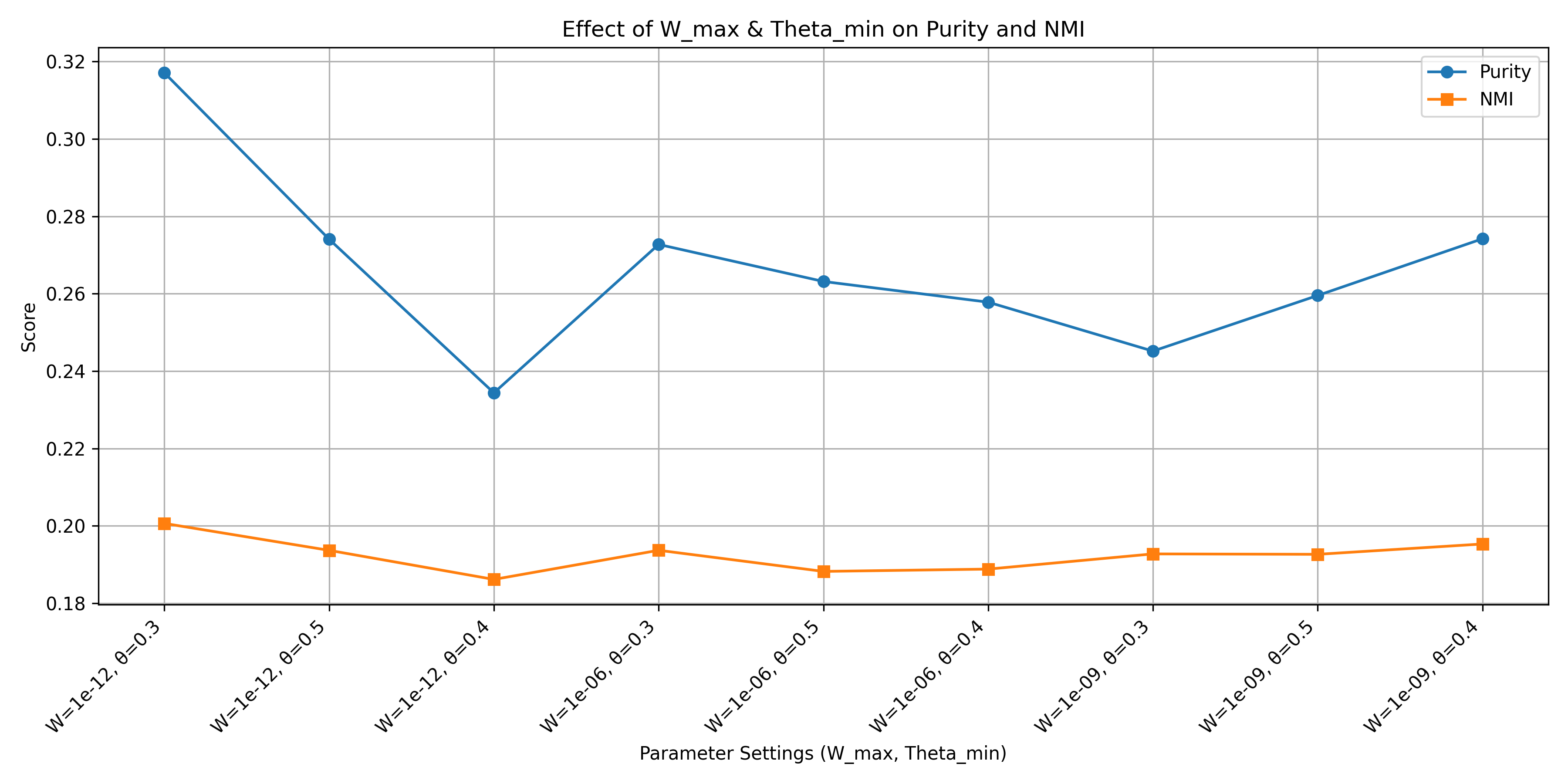}
\caption{Effect of \(W_{\text{max}}\) and \(\theta_{\text{min}}\) on NMI and purity scores.}
\label{fig:nmi_purity_sensitivity}
\end{figure*}

\begin{figure*}[!t]
\centering
\includegraphics[width=\linewidth]{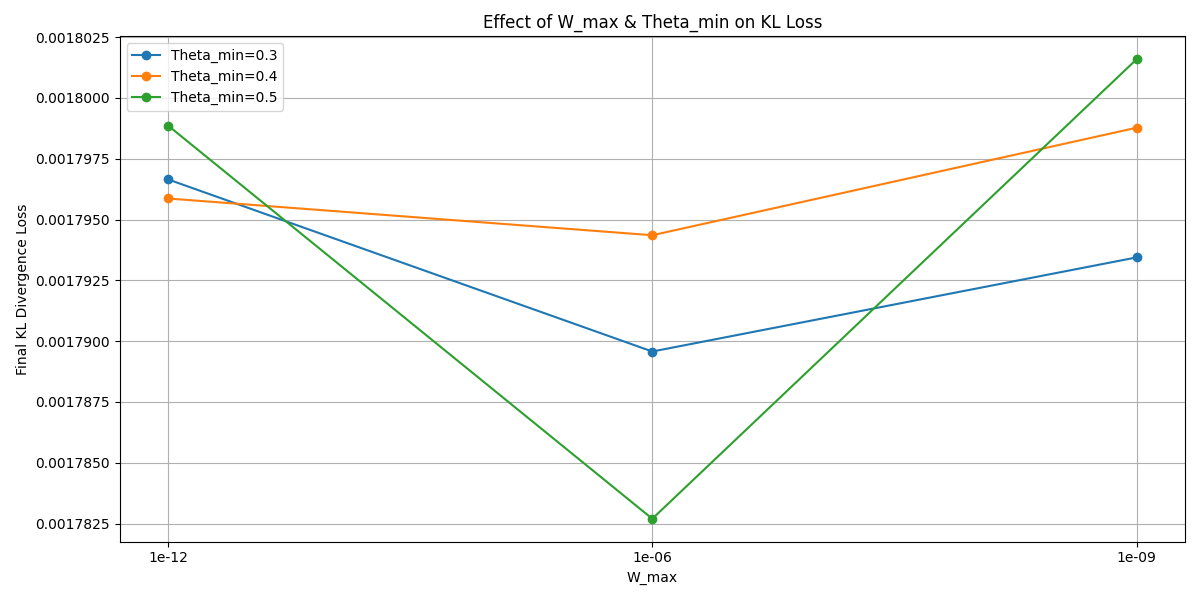}
\caption{Effect of \(W_{\text{max}}\) and \(\theta_{\text{min}}\) on KL divergence.}
\label{fig:kl_sensitivity}
\end{figure*}

We provide the ablation study referenced in the main text. These results highlight how the model responds to changes in key hyperparameters across various settings. As shown in Figures~\ref{fig:nmi_purity_sensitivity} and~\ref{fig:kl_sensitivity}, the model consistently performs well across different settings in terms of KL divergence, NMI, and purity.

\section{Additional Analysis on Varying Topic Counts and Topic Quality (JSD)}
\label{appendix:topic-count-analysis}
We further examined how changes in the number of total topics and minority topics affect evaluation metrics. As shown in Figures \ref{fig:nmi_scores(30,50,80)}, \ref{fig:purity_scores(30,50,80)}, our model consistently outperforms others across all configurations. Notably, it maintains strong performance even as the number of topics increases, which often challenges other models. We report results from synthetic dataset experiments using 30, 50, and 80 topics. In each setting, we designated 10, 15, and 20 topics, respectively, as minority (mental health-related) topics, with the remaining 20, 35, and 60 topics representing majority (non-mental health) themes. While most baselines show a drop in NMI and Purity, our approach increases, especially with larger sets of minority topics (e.g., K=80, MH=20).

Additionally, to assess topic coherence, we computed the JSD between topic distributions. Lower JSD indicates more distinct and well-formed topics. In Figure \ref{fig:JSD.png} (also shown in the main paper as Figure \ref{fig:jsd_evaluation_small}), curves are shown for the mean, standard deviation, minimum, and maximum of JSD over the ground-truth topics. Results show our model achieves high quality (small JS divergence); our model, NMF and KeyATM are the best three models outperforming others and are comparable to each other. Hence our model both attains high topic quality (small JS divergence) and outperforms all models in clustering ability (high purity \& NMI).

\begin{figure*}[!t]
\centering
\includegraphics[width=\linewidth]{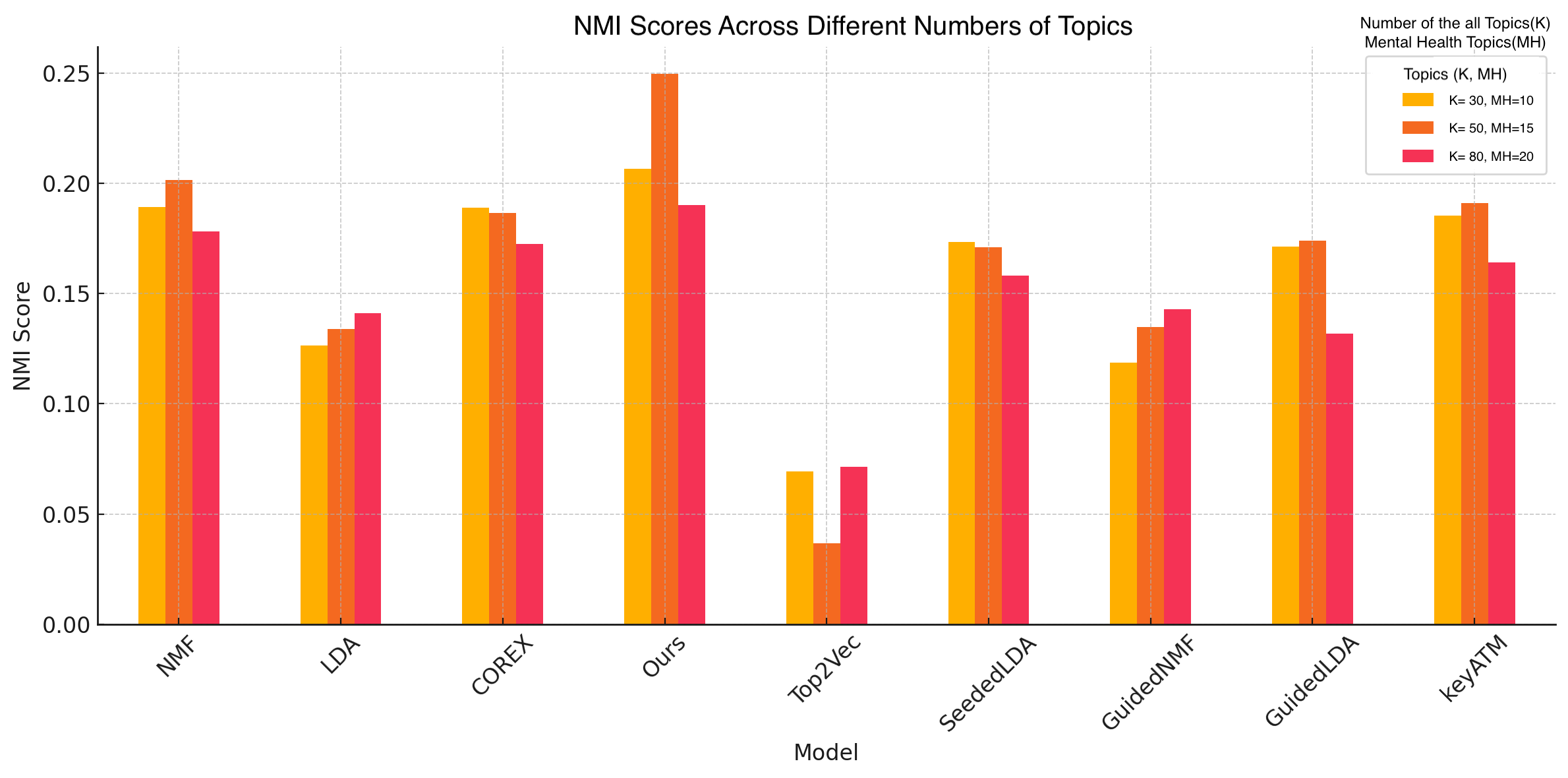}
\caption{NMI scores for topic counts 30, 50, and 80.}
\label{fig:nmi_scores(30,50,80)}
\end{figure*}

\begin{figure*}[!t]
\centering
\includegraphics[width=\textwidth]{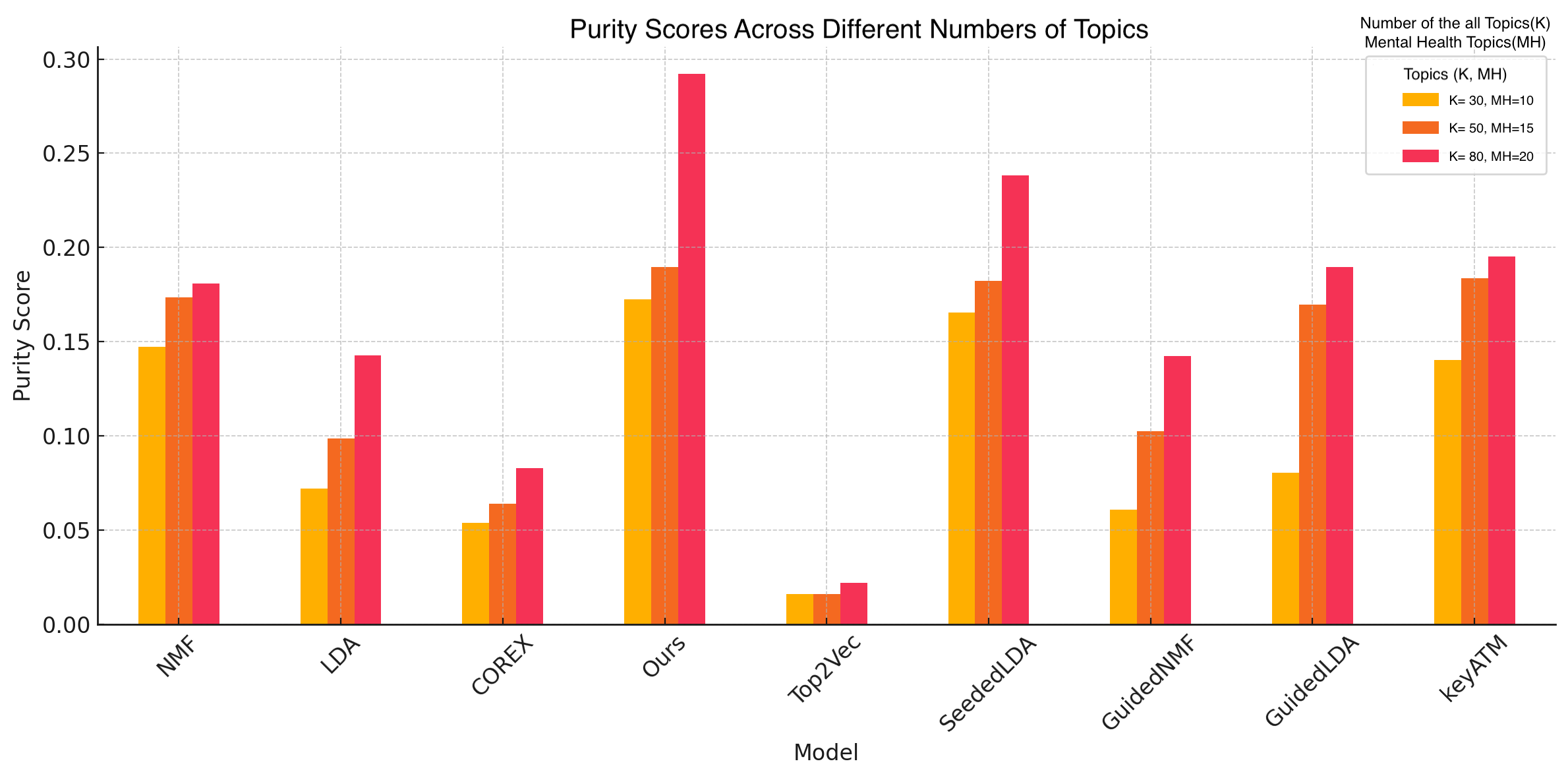}
\caption{Purity scores for topic counts 30, 50, and 80.}
\label{fig:purity_scores(30,50,80)}
\end{figure*}

\begin{figure*}[!t]
\centering
\includegraphics[width=\textwidth]{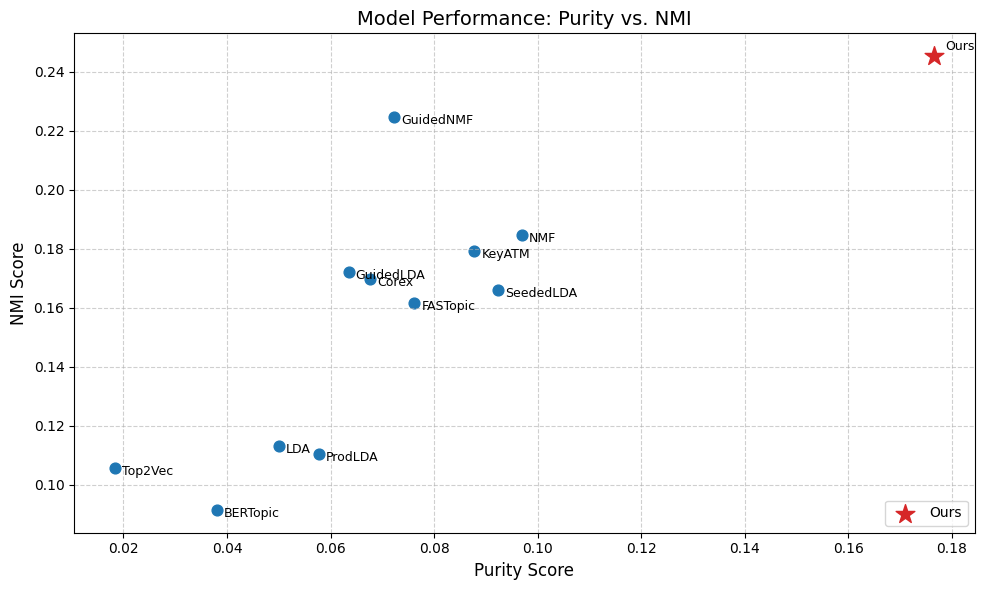}
\caption{Scatter plot of Purity vs. NMI scores for different models.}
\label{fig:purity_nmi_scatter}
\end{figure*}

\onecolumn


\end{document}